\documentclass[
10pt, 
a4paper, 
]{scrartcl}

\usepackage[affil-it]{authblk}
\usepackage{booktabs}
\usepackage{enumitem}
\usepackage{longtable}
\usepackage{graphicx}
\usepackage{enumerate}
\usepackage{array}
\usepackage{dcolumn}   
\usepackage{amsfonts} 
\usepackage{amssymb} 
\usepackage{xspace} 
\usepackage{algorithm}
\usepackage{algorithmic}
\usepackage{xfrac}
\usepackage{array} 
\usepackage{multirow}
\usepackage{xcolor}
\usepackage{footnote}
\usepackage{url}
\usepackage{makeidx}
\usepackage{epsfig}
\usepackage{subfig}
\usepackage{amsmath}
\usepackage{changes}
\usepackage{footnote}
\usepackage{rotating}
\usepackage{cite}
\usepackage[misc]{ifsym}
\begin{document}

\title{Evaluating time series forecasting models
}
\subtitle{An empirical study on performance estimation methods}



\author[1,2]{Vitor Cerqueira}
\author[1,2,3]{Luis Torgo}
\author[4]{Igor Mozeti\v{c}}
\affil[1]{LIAAD-INESC TEC, Porto, Portugal}
\affil[2]{University of Porto, Porto, Portugal}
\affil[3]{Dalhousie University, Halifax, Canada}
\affil[4]{Jo\v{z}ef Stefan Institute, Ljubljana, Slovenia}




\maketitle

\begin{abstract}
Performance estimation aims at estimating the loss that a predictive model will incur on unseen data. These procedures are part of the pipeline in every machine learning project and are used for assessing the overall generalisation ability of predictive models. In this paper we address the application of these methods to time series forecasting tasks. For independent and identically distributed data the most common approach is cross-validation. However, the dependency among observations in time series raises some caveats about the most appropriate way to estimate performance in this type of data and currently there is no settled way to do so. We compare different variants of cross-validation and  of out-of-sample approaches using two case studies: One with 62 real-world time series and another with three synthetic time series. Results show noticeable differences in the performance estimation methods in the two scenarios. In particular, empirical experiments suggest that cross-validation approaches can be applied to stationary time series. However, in real-world scenarios, when different sources of non-stationary variation are at play, the most accurate estimates are produced by out-of-sample methods that preserve the temporal order of observations.
\end{abstract}

\section{Introduction}\label{intro}

Machine learning plays an increasingly important role in science and technology, and performance estimation is part of any machine learning project pipeline. This task is related to the process of using the available data to estimate the loss that a model will incur on unseen data. Machine learning practitioners typically use these methods for model selection, hyper-parameter tuning and assessing the overall generalization ability of the models. In effect, obtaining reliable estimates of the performance of models is a critical issue on all predictive analytics tasks.

Choosing a performance estimation method often depends on the data one is modelling. For example, when one can assume independence and an identical distribution (i.i.d.) among observations, cross-validation~\cite{geisser1975predictive} is typically the most appropriate method. This is mainly due to its efficient use of data~\cite{arlot2010survey}. 
However, there are some issues when the observations in the data are dependent, such as time series. These dependencies raise some caveats about using standard cross-validation in such data. Notwithstanding, there are particular time series settings in which variants of cross-validation can be used, such as in stationary or small-sized data sets where the efficient use of all the data by cross-validation is beneficial~\cite{bergmeir2015note}.

In this paper we present a comparative study of different performance estimation methods for time series forecasting tasks. Several strategies have been proposed in the literature and currently there is no consensual approach. We applied different methods in two case studies. One is comprised of 62 real-world time series with potential non-stationarities and the other is a stationary synthetic environment~\cite{bergmeir2012use,bergmeir2014usefulness,bergmeir2015note}. 

In this study we compare two main classes of estimation methods: 

\begin{itemize}
    \item Out-of-sample (OOS): These methods have been traditionally used to estimate predictive performance in time-dependent data. Essentially, out-of-sample methods hold out the last part of the time series for testing. Although these approaches do not make a complete use of the available data, they preserve the temporal order of observations. This property may be important to cope with  the dependency among observations and account for the potential temporal correlation between the consecutive values of the time series.
    \item Cross-validation (CVAL): These approaches make a more efficient use of the available data, which is beneficial in some settings~\cite{bergmeir2015note}. They assume that observations are i.i.d., though some strategies have been proposed to circumvent this requirement. These methods have been shown to be able to provide more robust estimations than out-of-sample approaches in some time series scenarios~\cite{bergmeir2012use,bergmeir2014usefulness,bergmeir2015note}.  
\end{itemize}

\noindent A key characteristic that distinguishes these two types of approaches is that OOS methods always preserve the temporal order of observations meaning that a model is never tested on past data. The objective of this study is to address the following research question: How do out-of-sample methods compare to cross-validation approaches in terms of performance estimation ability for different types of time series data? 

This paper is an extension to an article published before \cite{cerqueira2017comparative}. In this work, we substantially increase the experimental setup both in methods and data sets used; provide additional analysis such as the impact of stationarity; and a more in-depth and critical discussion of the results.

This paper is structured as follows. The literature on performance estimation for time series forecasting tasks is reviewed in Section~\ref{sec:lr}. Materials and methods are described in Section~\ref{sec:materialsmethods}, including the predictive task, time series data sets, performance estimation methodology, and experimental design. The results of the experiments are reported in Section~\ref{sec:ee}. A  discussion of our results is carried out in Section~\ref{sec:disc}. Finally, the conclusions of our empirical study are provided in Section~\ref{sec:fr}.

\section{Background}\label{sec:lr}

In this section we provide a background to this paper. We review the typical estimation methods used in time series forecasting and explain the motivation for this study.

In general, performance estimation methods for time series forecasting tasks are designed to cope with the dependence between observations. This is typically accomplished by having a model tested on observations future to the ones used for training. These include the OOS testing as well as variants of the CVAL method.

\subsection{Out-of-sample approaches}

\begin{figure}[thb]
    \centering
    \includegraphics[width=.8\textwidth, trim=0cm 0cm 0cm 0cm, clip=TRUE]{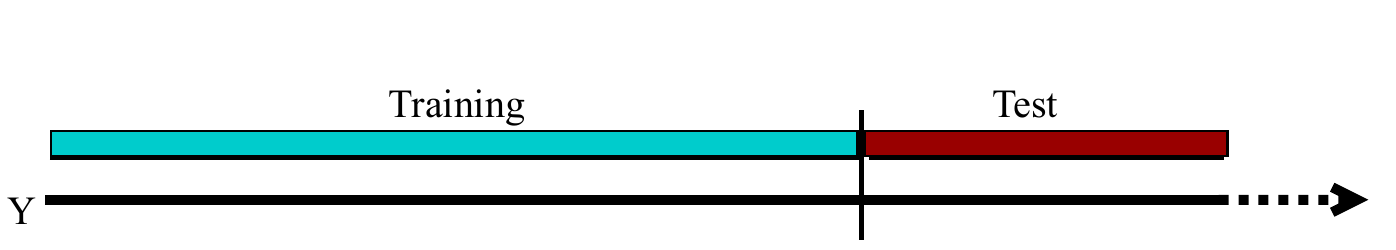}
    \caption{Simple out-of-sample procedure: an initial part of the available observations are used for fitting a predictive model. The last part of the data is held out, where the predictive model is tested.}
    \label{fig:houtapproaches}
\end{figure}

\begin{figure}[bth]
    \centering
    \includegraphics[width=.8\textwidth, trim=0cm 0cm 0cm 0cm, clip=TRUE]{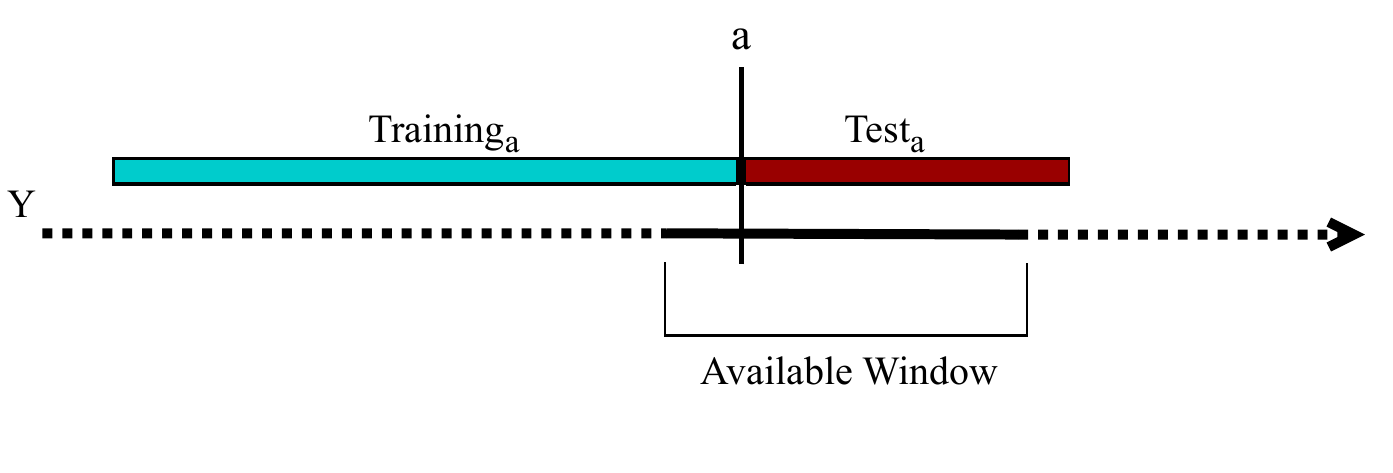}
    \caption{Example of one iteration of the repeated holdout procedure. A point $a$ is chosen from the available window. Then, a previous part of observations are used for training, while a subsequent part of observations are used for testing.}
    \label{fig:rephoutapproaches}
\end{figure}

When using OOS performance estimation procedures, a time series is split into two parts: an initial fit period in which a model is trained, and a testing period held out for estimating the loss of that model. This simple approach (\texttt{Holdout}) is depicted in Figure~\ref{fig:houtapproaches}. However, within this type of procedure one can adopt different strategies regarding training/testing split point, growing or sliding window settings, and eventual update of the models. In order to produce a robust estimate of predictive performance, Tashman~\cite{tashman2000out} recommends employing these strategies in multiple test periods. One might create different sub-samples according to, for example, business cycles~\cite{fildes1989evaluation}. For a more general setting one can also adopt a randomized approach. This is similar to random sub-sampling (or repeated holdout) in the sense that they consist of repeating a learning plus testing cycle several times using different, but possibly overlapping data samples (\texttt{Rep-Holdout}). This idea is illustrated in Figure~\ref{fig:rephoutapproaches}, where one iteration of a repeated holdout is shown. A point $a$ is randomly chosen from the available window (constrained by the training and testing sizes) of a time series Y. This point then marks the end of the training set, and the start of the testing set.

\subsection{Prequential}

\begin{figure}[bth]
    \centering
    \includegraphics[width=.8\textwidth, trim=0cm 0cm 0cm 0cm, clip=TRUE]{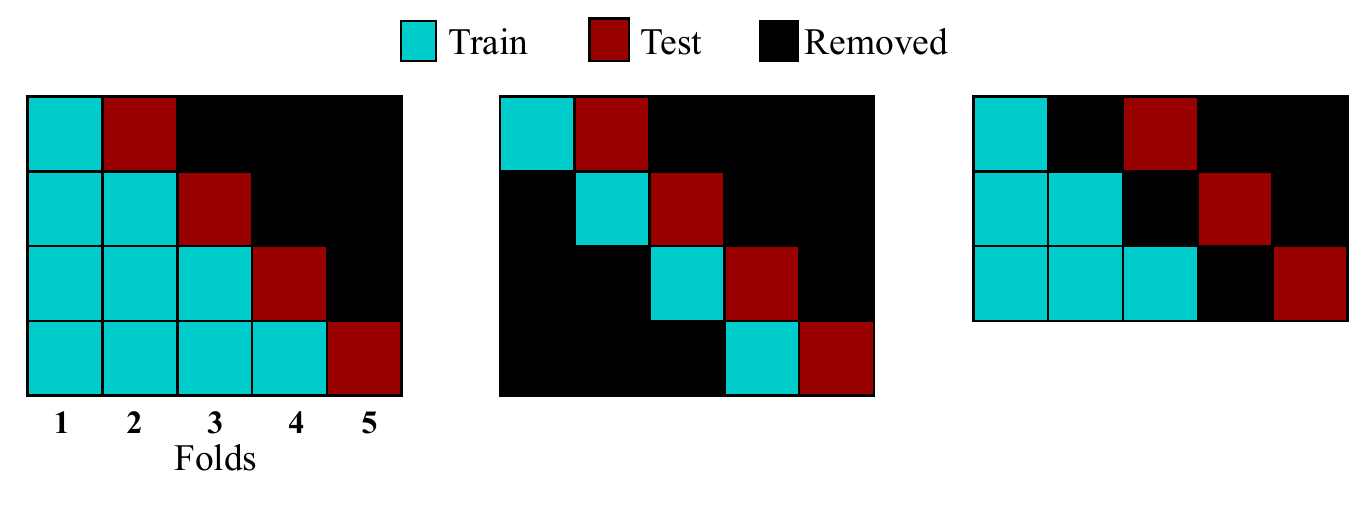}
    \caption{Variants of prequential approach applied in blocks for performance estimation. This strategy can be applied using a growing window (left, right), or a sliding window (middle). One can also introduce a gap between the training and test sets.}
    \label{fig:preqapproaches}
\end{figure}

OOS approaches are similar to prequential or interleaved-test-then-train evaluation~\cite[Chapter~2.2]{bifet2009data}. Prequential is typically used in data streams mining. The idea is that each observation is first used to test the model, and then to train the model. This can be applied in blocks of sequential instances~\cite{modha1998prequential}. In the initial iteration, only the first two blocks are used, the first for training and the second for test. In the next iteration, the second block is merged with the first and the third block is used for test. This procedure continues until all blocks are tested (\texttt{Preq-Bls}). This procedure is exemplified in the left side of Figure~\ref{fig:preqapproaches}, in which the data is split into 5 blocks.

A variant of this idea is illustrated in the middle scheme of Figure~\ref{fig:preqapproaches}. Instead of merging the blocks after each iteration (growing window), one can forget the older blocks in a sliding window fashion (\texttt{Preq-Sld-Bls}). This idea is typically adopted when past data becomes deprecated, which is common in non-stationary environments. Another variant of the prequential approach is represented in the right side of Figure~\ref{fig:preqapproaches}. This illustrates a prequential approach applied in blocks, where a gap block is introduced (\texttt{Preq-Bls-Gap}). The rationale behind this idea is to increase the independence between training and test sets.

\subsection{Cross-validation approaches}

The typical approach when using K-fold cross-validation is to randomly shuffle the data and split it in K equally-sized folds or blocks. Each fold is a subset of the data comprising $t/K$ randomly assigned observations, where $t$ is the number of observations. After splitting the data into K folds, each fold is iteratively picked for testing. A model is trained on K-1 folds and its loss is estimated on the left out fold (\texttt{CV}). In fact, the initial random shuffle of observations before splitting into different blocks is not intrinsic to cross-validation~\cite{geisser1975predictive}. Notwithstanding, the random shuffling is a common practice among data science professionals. This approach to cross-validation is illustrated in the left side of Figure~\ref{fig:cvapproaches}.

\subsubsection{Variants designed for time-dependent data}

Some variants of K-fold cross-validation have been proposed specially designed for dependent data, such as time series~\cite{arlot2010survey}. However, theoretical problems arise by applying this technique directly to this type of data. The dependency among observations is not taken into account since cross-validation assumes the observations to be i.i.d.. This might lead to overly optimistic estimations and consequently, poor generalisation ability of predictive models on new observations. For example, prior work has shown that cross-validation yields poor estimations for the task of choosing the bandwidth of a kernel estimator in correlated data~\cite{hart1986kernel}. To overcome this issue and approximate independence between the training and test sets, several methods have been proposed as variants of this procedure. We will focus on variants designed to cope with temporal dependency among observations.

\begin{figure}[th]
    \centering
    \includegraphics[width=.9\textwidth, trim=0cm 0cm 0cm 0cm, clip=TRUE]{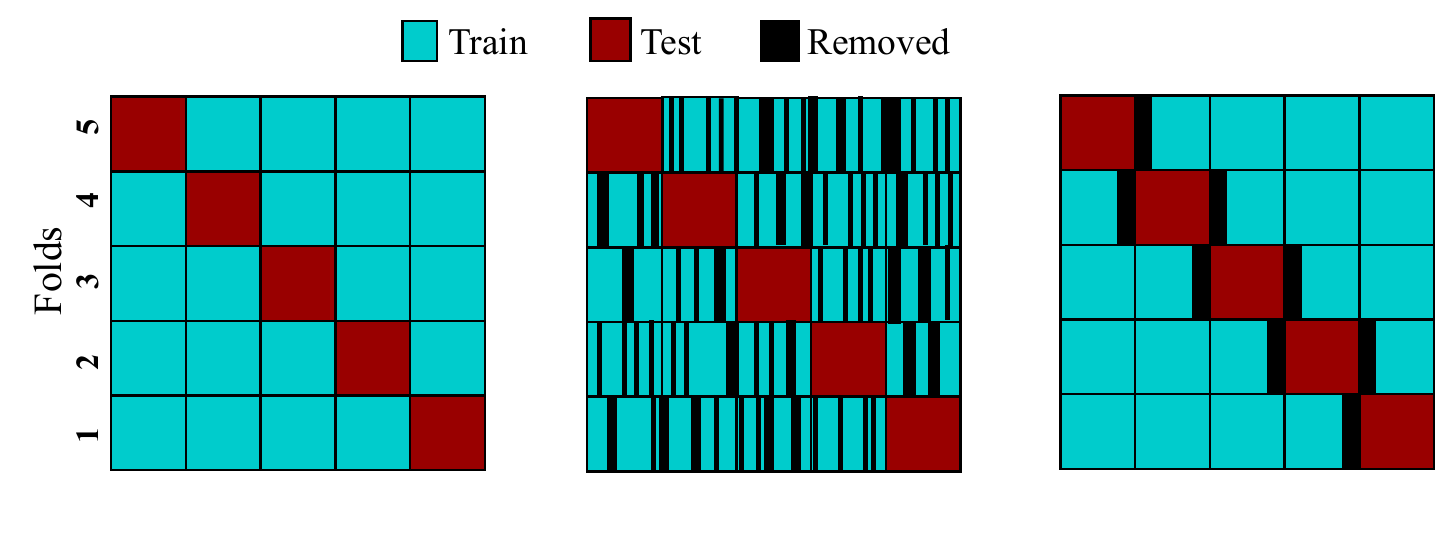}
    \caption{Variants of cross-validation estimation procedures}
    \label{fig:cvapproaches}
\end{figure}

The Blocked Cross-Validation~\cite{snijders1988cross} (\texttt{CV-Bl}) procedure is similar to the standard form described above. The difference is that there is no initial random shuffling of observations. In time series, this renders $K$ blocks of contiguous observations. The natural order of observations is kept within each block, but broken across them. This approach to cross-validation is also illustrated in the left side of Figure~\ref{fig:cvapproaches}. Since the random shuffle of observations is not being illustrated, the figure for \texttt{CV-Bl} is identical to the one shown for \texttt{CV}. 

The Modified CV procedure~\cite{mcquarrie1998regression} (\texttt{CV-Mod}) works by removing observations from the training set that are correlated with the test set. The data is initially randomly shuffled and split into $K$ equally-sized folds similarly to K-fold cross-validation. Afterwards, observations from the training set within a certain temporal range of the observations of the test set are removed. This ensures independence between the training and test sets. However, when a significant amount of observations are removed from training, this may lead to model under-fit. This approach is also described as non-dependent cross-validation~\cite{bergmeir2012use}. The graph in the middle of Figure~\ref{fig:cvapproaches} illustrates this approach.

The hv-Blocked Cross-Validation (\texttt{CV-hvBl}) proposed by Racine~\cite{racine2000consistent} extends blocked cross-validation to further increase the independence among observations. Specifically, besides blocking the observations in each fold, which means there is no initial randomly shuffle of observations, it also removes adjacent observations between the training and test sets. Effectively, this creates a gap between both sets. This idea is depicted in the right side of Figure~\ref{fig:cvapproaches}.

\subsubsection{Usefulness of cross-validation approaches}

Recently there has been some work on the usefulness of cross-validation procedures for time series forecasting tasks. Bergmeir and Ben\'{i}tez~\cite{bergmeir2012use} present a comparative study of estimation procedures using stationary time series. Their empirical results show evidence that in such conditions cross-validation procedures yield more accurate estimates than an OOS approach. Despite the theoretical issue of applying standard cross-validation, they found no practical problem in their experiments. Notwithstanding, the Blocked cross-validation is suggested for performance estimation using stationary time series. 

Bergmeir et al.~\cite{bergmeir2014usefulness} extended their previous work for directional time series forecasting tasks. These tasks are related to predicting the direction (upward or downward) of the observable. The results from their experiments suggest that the hv-Blocked CV procedure provides more accurate estimates than the standard out-of-sample approach. These were obtained by applying the methods on stationary time series.

Finally, Bergmeir et al.~\cite{bergmeir2015note} present a simulation study comparing standard cross-validation to out-of-sample evaluation. They used three data generating processes and performed 1000 Monte Carlo trials in each of them. For each trial and generating process, a stationary time series with 200 values was created. The results from the simulation suggest that cross-validation systematically yields more accurate estimates, provided that the model is correctly specified.

In a related empirical study \cite{mozetic2018evaluate}, the authors compare estimation procedures on several large time-ordered Twitter datasets. They find no significant difference between the best cross-validation and out-of-sample evaluation procedures. However, they do find that standard, randomized  cross-validation is significantly worse than the blocked cross-validation, and should not be used to evaluate classifiers in time-ordered data scenarios.

Despite the results provided by these previous works we argue that they are limited in two ways. First, the used experimental procedure is biased towards cross-validation approaches. While these produce several error estimates (one for each fold), the OOS approach is evaluated in a one-shot estimation, where the last part of the time series is withheld for testing. OOS methods can be applied in several windows for more robust estimates, as recommended by Tashman~\cite{tashman2000out}. By using a single origin, one is prone to particular issues related to that origin.

Second, the results are based on stationary time series, most of them artificial. Time series stationarity is equivalent to identical distribution in the terminology of more traditional predictive tasks. Hence, the synthetic data generation processes and especially the stationary assumption limit interesting patterns that can occur in real-world time series. Our working hypothesis is that in more realistic scenarios one is likely to find time series with complex sources of non-stationary variations.

In this context, this paper provides an extensive comparative study using a wide set of methods for evaluating the performance of uni-variate time series forecasting models. These include several variants of both cross-validation and out-of-sample approaches. The analysis is carried out using a real-world scenario as well as a synthetic case study used in the works described previously~\cite{bergmeir2012use,bergmeir2014usefulness,bergmeir2015note}.

\subsection{Related work on performance estimation for dependent data}

The problem of performance estimation has also been under research in different scenarios where the observations are somehow dependent (non-i.i.d.).

\subsubsection{Performance estimation under spatio-temporal dependencies}

Geo-referenced time series are becoming more prevalent due to the increase of data collection from sensor networks. In these scenarios, the most appropriate estimation procedure is not obvious as spatio-temporal dependencies are at play. Oliveira et al.~\cite{oliveira2018evaluation} presented an extensive empirical study of performance estimation for forecasting problems with spatio-temporal time series. The results reported by the authors suggest that both CVAL and OOS methods are applicable in these scenarios. Like previous work in time-dependent domains \cite{bergmeir2012use,mozetic2018evaluate}, Oliveira et al. suggest the use of blocking when using a cross-validation estimation procedure.

\subsubsection{Performance estimation in data streams mining}

Data streams mining is concerned with predictive models that evolve continuously over time in response to concept drift~\cite{gama2014survey}. Gama et al.~\cite{gama2013evaluating} provide a thorough overview of the evaluation of predictive models for data streams mining. The authors defend the usage of the prequential estimator with a forgetting mechanism, such as a fading factor or a sliding window. 

This work is related to ours in the sense that it deals with performance estimation using time-dependent data. Notwithstanding, the paradigm of data streams mining is in line with sequential analysis~\cite{wald1973sequential}. As such, the assumption is that the sample size is not fixed in advance, and predictive models are evaluated as observations are collected. In our setting, given a time series data set, we want to estimate the loss that a predictive models will incur in unseen observations future to that data set.

\section{Materials and methods}\label{sec:materialsmethods}

In this section we present the materials and methods used in this work. First, we will define the prediction task. Second, the time series data sets are described. We then formalize the methodology employed for performance estimation. Finally, we overview the experimental design.

\subsection{Predictive task definition}

A time series is a temporal sequence of values $Y = \{y_1, y_2, \dots, y_t \}$, where $y_i$ is the value of $Y$ at time $i$ and $t$ is the length of $Y$. We remark that we use the term time series assuming that $Y$ is a numeric variable, i.e., $y_i \in \mathbb{R}, \forall$ $y_i \in Y$.

Time series forecasting denotes the task of predicting the next value of the time series, $y_{t+1}$, given the previous observations of $Y$. We focus on a purely auto-regressive modelling approach, predicting future values of time series using its past lags. 

To be more precise, we use time delay embedding~\cite{Takens1981} to represent $Y$ in an Euclidean space with embedding dimension $p$. Effectively, we construct a set of observations which are based on the past $p$ lags of the time series. Each observation is composed of a feature vector $x_i \in \mathbb{X} \subset \mathbb{R}^p$, which denotes the previous $p$ values, and a target vector $y_i \in \mathbb{Y} \subset \mathbb{R}$, which represents the value we want to predict. The objective is to construct a model $f : \mathbb{X} \rightarrow \mathbb{Y}$, where $f$ denotes the regression function.

Summarizing, we generate the following matrix:

\[
    Y_{[n,p]} = \left[
    \begin{array}{ccccc|c}
    y_{1} & y_{2} & \dots  & y_{p-1} & y_{p} & y_{p+1} \\ 

    \vdots & \vdots & \vdots  & \vdots & \vdots & \vdots\\
    
    y_{i-p+1} & y_{i-p+2} & \dots  & y_{i-1} & y_{i} & y_{i+1}\\
    
    \vdots & \vdots & \vdots  & \vdots & \vdots & \vdots\\
    
    y_{t-p+1} & y_{t-p+2} & \dots  & y_{t-1} & y_{t} & y_{t+1}
    \end{array}
        \right]
    \]

Taking the first row of the matrix as an example, the target value is $y_{p+1}$, while the attributes (predictors) are the previous $p$ values $\{y_1, \dots, y_{p}\}$. Essentially we assume that there are no time dependencies larger than $p$. 

\subsection{Time series data}\label{sec:cs}

Two different case studies are used to analyse the performance estimation methods: a scenario comprised of real-world time series and a synthetic setting used in prior work~\cite{bergmeir2012use,bergmeir2014usefulness,bergmeir2015note} for addressing the issue of performance estimation for time series forecasting tasks. 

\subsubsection{Real-world time series}\label{sec:stationarity}

We analyse 62 real-world time series (RWTS) from different domains. They have different granularity and length as well as unknown dynamics. The time series are described in Table~\ref{tab:data} in Appendix \ref{appendix}. In the table, the column $p$ denotes the embedding dimension of the respective time series. Our approach for estimating this parameter is addressed in section \ref{sec:embeddingdim}. Differencing is the computation of the differences between consecutive observations. This process is useful to remove changes in the level of a time series, thus stabilising the mean \cite{hyndman2018forecasting}. This is important to account for trend and seasonality in time series. The column I represents the number of differences applied to the respective time series in order to make it trend-stationary according to the KPSS test~\cite{kwiatkowski1992testing}. Finally, the column S represents whether or not a time series is stationary (1 if it is, 0 otherwise).

We analysed the stationarity of the time series comprising the real-world case study. Essentially, a time series is said to be stationary if its characteristics do not depend on the time that the data is observed~\cite{hyndman2018forecasting}. In this work we consider a stationarity of order 2. This means that a time series is considered stationary if it has constant mean, constant variance, and an auto-covariance that does not depend on time. Henceforth we will refer a time series as stationary if it is stationary of order 2. 

In order to test if a given time series is stationary we follow the wavelet spectrum test described by Nason~\cite{nason2013test}. This test starts by computing an evolutionary wavelet spectral approximation. Then, for each scale of this approximation, the coefficients of the Haar wavelet are computed. Any large Haar coefficient is evidence of a non-stationarity. An hypothesis test is carried out to assess if a coefficient is large enough to reject the null hypothesis of stationarity. In particular, we apply a multiple hypothesis test with a Bonferroni correction and a false discovery rate~\cite{nason2013test}.

\begin{figure}[ht]
    \centering
    \includegraphics[width=.8\textwidth, trim=1cm .5cm 0cm 0cm, clip=TRUE]{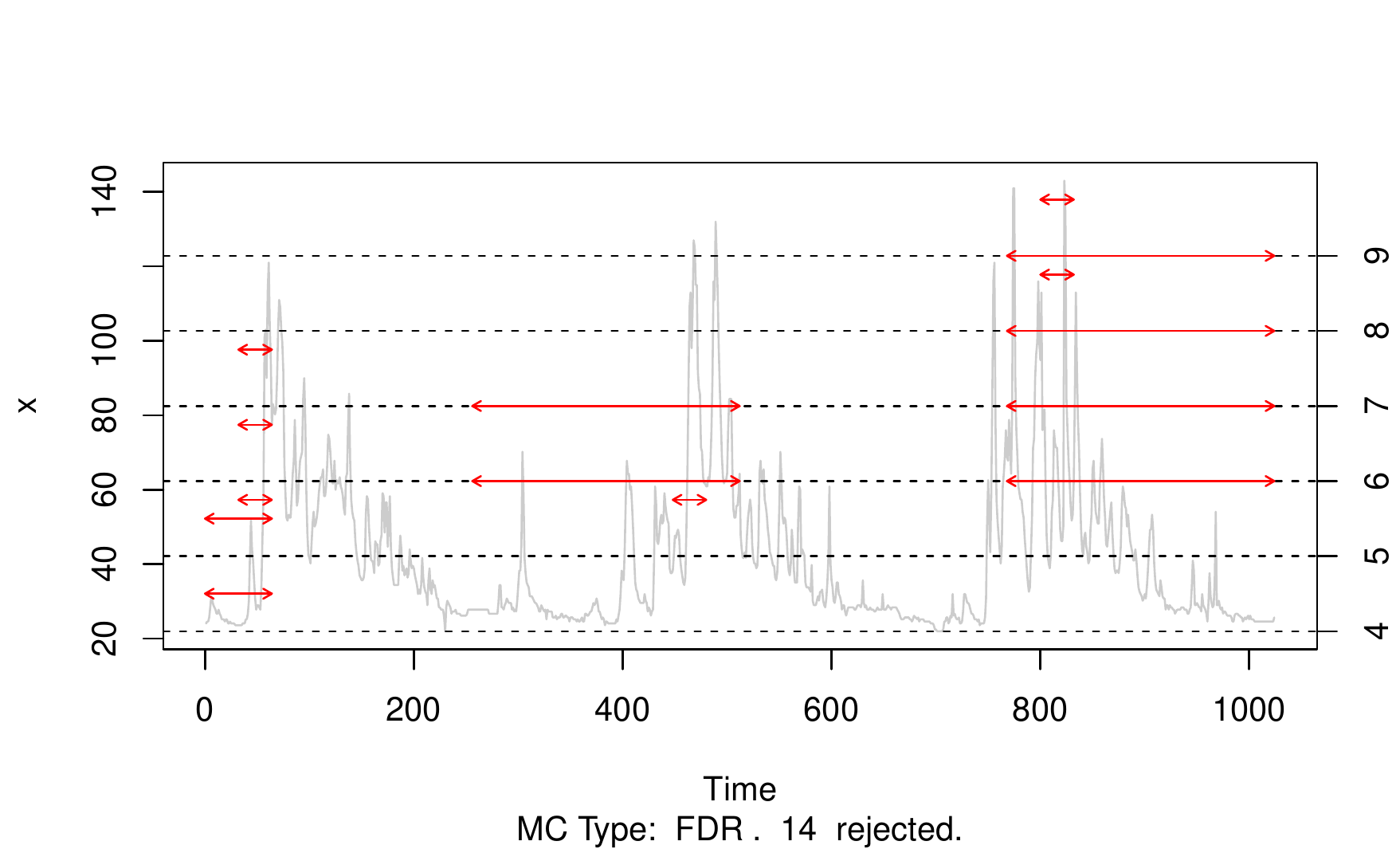}
    \caption{Application of the wavelet spectrum test to a non-stationary time series. Each red horizontal arrow denote a non-stationarity identified by the test.}
    \label{fig:eg_nonst}
\end{figure}

In Figure~\ref{fig:eg_nonst} is shown an example of the application of the wavelet spectrum test to a non-stationary time series. In the graphic, each red horizontal arrow denotes a non-stationarity found by the test. The left-hand side axis denotes the scale of the time series. The right-hand axis represents the scale of the wavelet periodogram and where the non-stationarities are found. Finally, the lengths of the arrows denote the scale of the Haar wavelet coefficient whose null hypothesis was rejected. For a thorough description of this method we refer to the work by Nason~\cite{nason2013test}.

\subsubsection{Synthetic time series}

\begin{figure}[ht]
    \centering
    \includegraphics[width=.8\textwidth, trim=0cm 0cm 0cm 0cm, clip=TRUE]{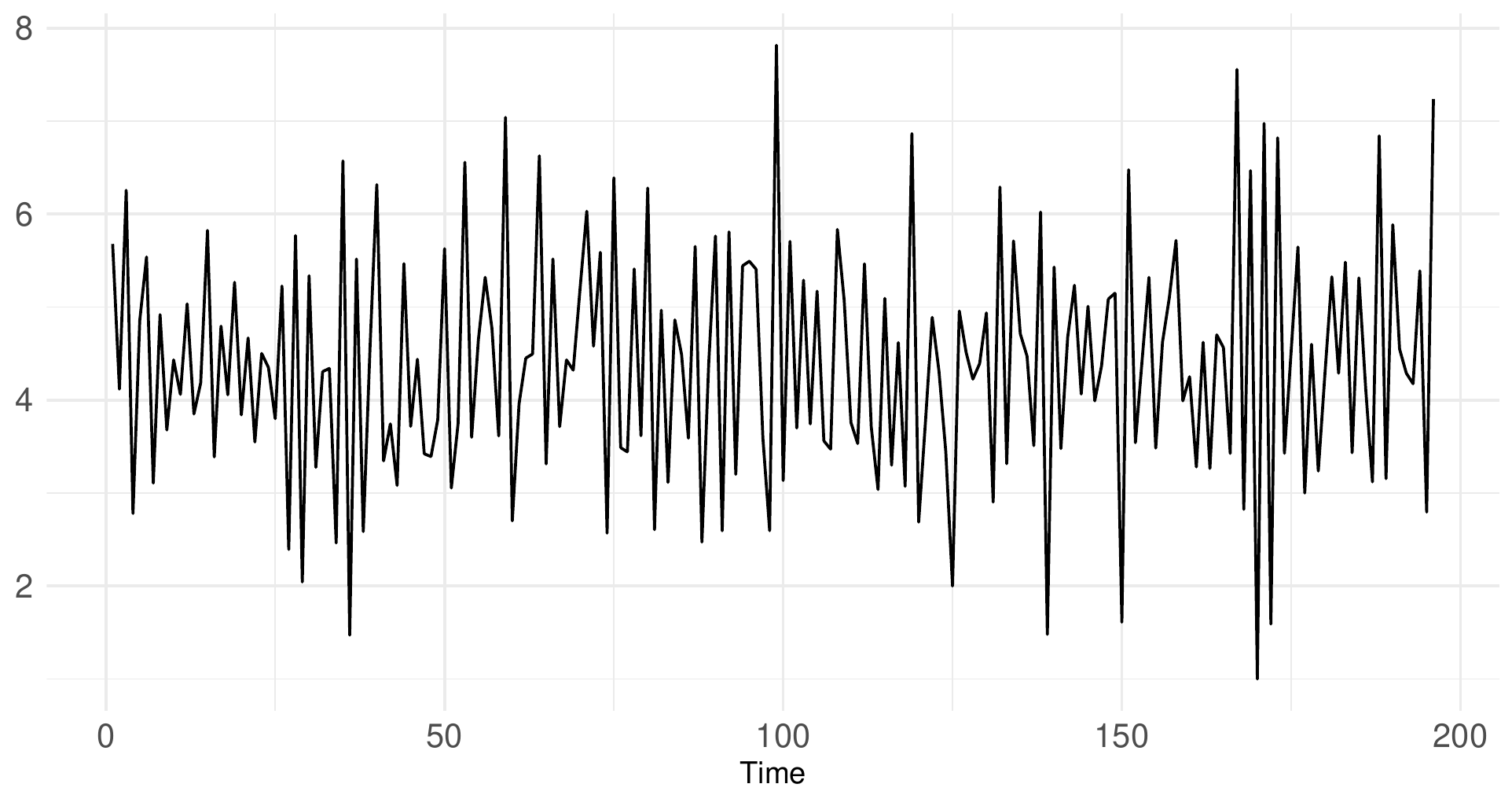}
    \caption{Sample graph of the S1 synthetic case.}
    \label{fig:s1_plot}
\end{figure}

\begin{figure}[ht]
    \centering
    \includegraphics[width=.8\textwidth, trim=0cm 0cm 0cm 0cm, clip=TRUE]{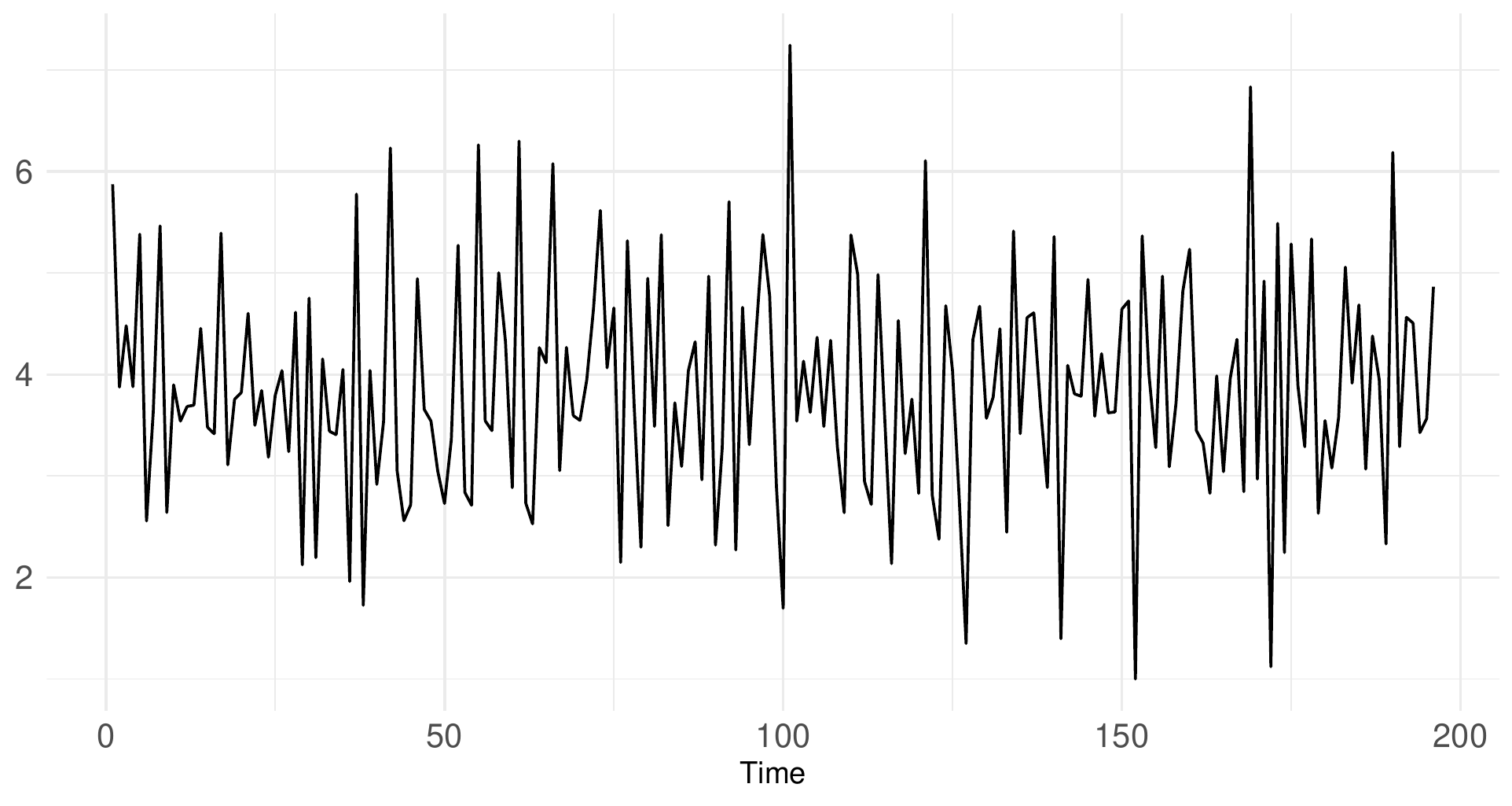}
    \caption{Sample graph of the S2 synthetic case.}
    \label{fig:s2_plot}
\end{figure}

\begin{figure}[ht]
    \centering
    \includegraphics[width=.8\textwidth, trim=0cm 0cm 0cm 0cm, clip=TRUE]{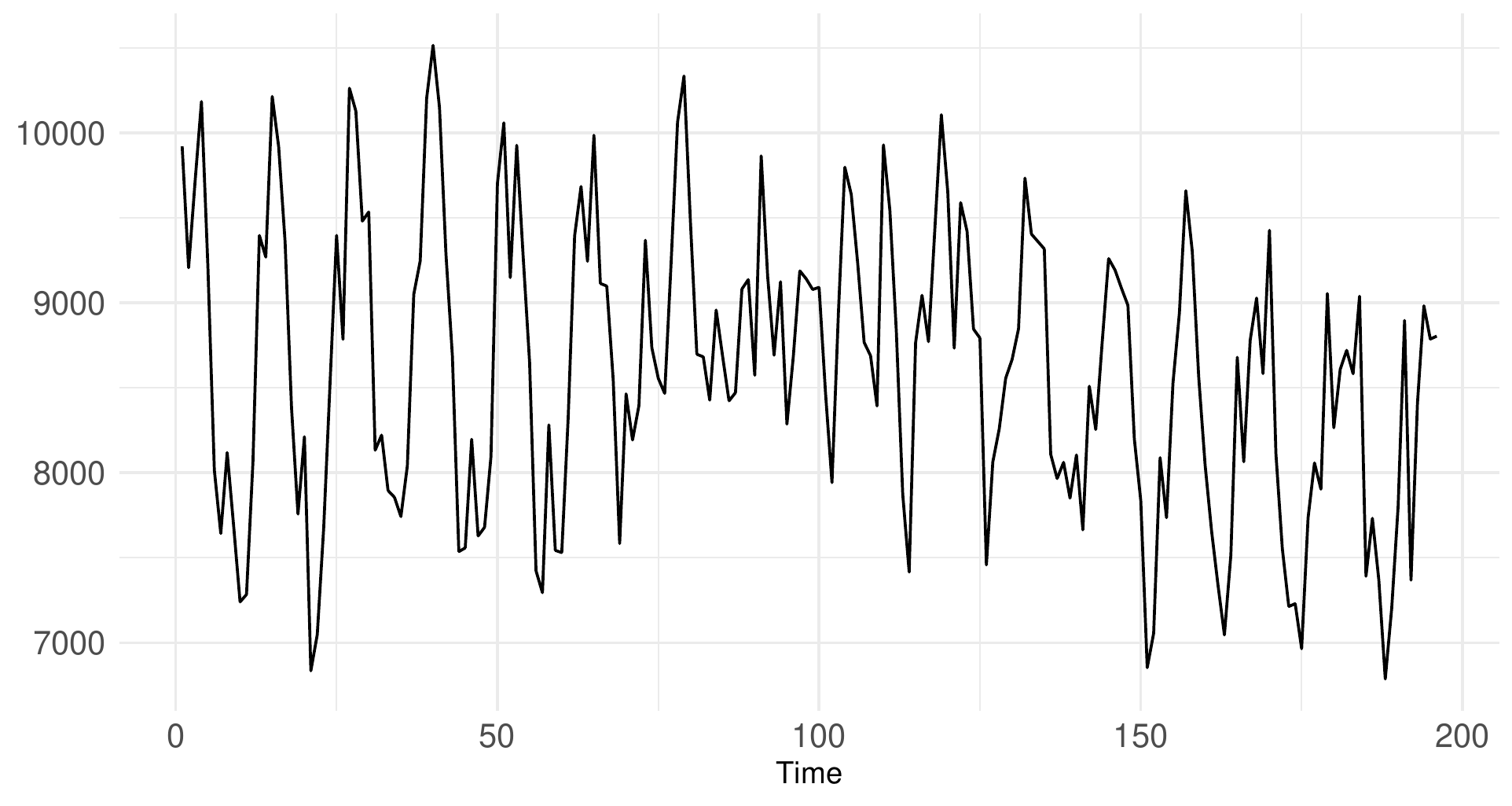}
    \caption{Sample graph of the S3 synthetic case.}
    \label{fig:s3_plot}
\end{figure}

We use three synthetic use cases defined in previous work by Bergmeir et al.~\cite{bergmeir2014usefulness,bergmeir2015note}. The data generating processes are all stationary and are designed as follows: 

\begin{description}[leftmargin=*]
	\item[\textbf{S1:}] A stable auto-regressive process with lag 3, i.e., the next value of the time series is  dependent on the past 3 observations -- c.f. Figure~\ref{fig:s1_plot} for a sample graph.
	\item[\textbf{S2}:] An invertible moving average process with lag 1 -- c.f. Figure~\ref{fig:s2_plot} for a sample graph.
	\item[\textbf{S3}:] A seasonal auto-regressive process with lag 12 and seasonal lag 1 -- c.f. Figure~\ref{fig:s3_plot} for a sample graph.
\end{description}

For the first two cases, S1 and S2, real-valued roots of the characteristic polynomial are sampled from the uniform distribution $[-r;-1.1]\cup[1.1,r]$, where $r$ is set to 5~\cite{bergmeir2012use}. Afterwards, the roots are used to estimate the models and create the time series. The data is then processed by making the values all positive. This is accomplished by subtracting the minimum value and adding 1. The third case S3 is created by fitting a seasonal auto-regressive model to a time series of monthly total accidental deaths in the USA~\cite{brockwell2013time}. For a complete description of the data generating process we refer to the work by Bergmeir et al.~\cite{bergmeir2012use,bergmeir2015note}. Similarly to Bergmeir et al., for each use case we performed 1000 Monte Carlo simulations. In each repetition a time series with 200 values was generated.

\subsection{Performance estimation methodology}

Performance estimation addresses the issue of estimating the predictive performance of predictive models. Frequently, the objective behind these tasks is to compare different solutions for solving a predictive task. This includes selecting among different learning algorithms and hyper-parameter tuning for a particular one.

Training a learning model and evaluating its predictive ability on the same data has been proven to produce biased results due to overfitting~\cite{arlot2010survey}. Since then several methods for performance estimation have been proposed in the literature, which use new data to estimate the performance of models. Usually, new data is simulated by splitting the available data. Part of the data is used for training the learning algorithm and the remaining data is used to test and estimate the performance of the model.

For many predictive tasks the most widely used of these methods is K-fold cross-validation~\cite{stone1974cross} (c.f. Section~\ref{sec:lr} for a description). The main advantages of this method is its universal splitting criteria and efficient use of all the data. However, cross-validation is based on the assumption that observations in the underlying data are independent. When this assumption is violated, for example in time series data, theoretical problems arise that prevent the proper use of this method in such scenarios. As we described in Section~\ref{sec:lr} several methods have been developed to cope with this issue, from out-of-sample approaches~\cite{tashman2000out} to variants of the standard cross-validation, e.g., block cross-validation~\cite{snijders1988cross}. 

Our goal in this paper is to compare a wide set of estimation procedures, and test their suitability for different types of time series forecasting tasks. In order to emulate a realistic scenario we split each time series data in two parts. The first part is used to estimate the loss that a given learning model will incur on unseen future observations. This part is further split into training and test sets as described before. The second part is used to compute the true loss that the model incurred. This strategy allows the computation of unbiased estimates of error since a model is always tested on unseen observations.

\begin{figure}[ht]
    \centering
    \includegraphics[width=.6\textwidth, trim=0cm 0cm 0cm 0cm, clip=TRUE]{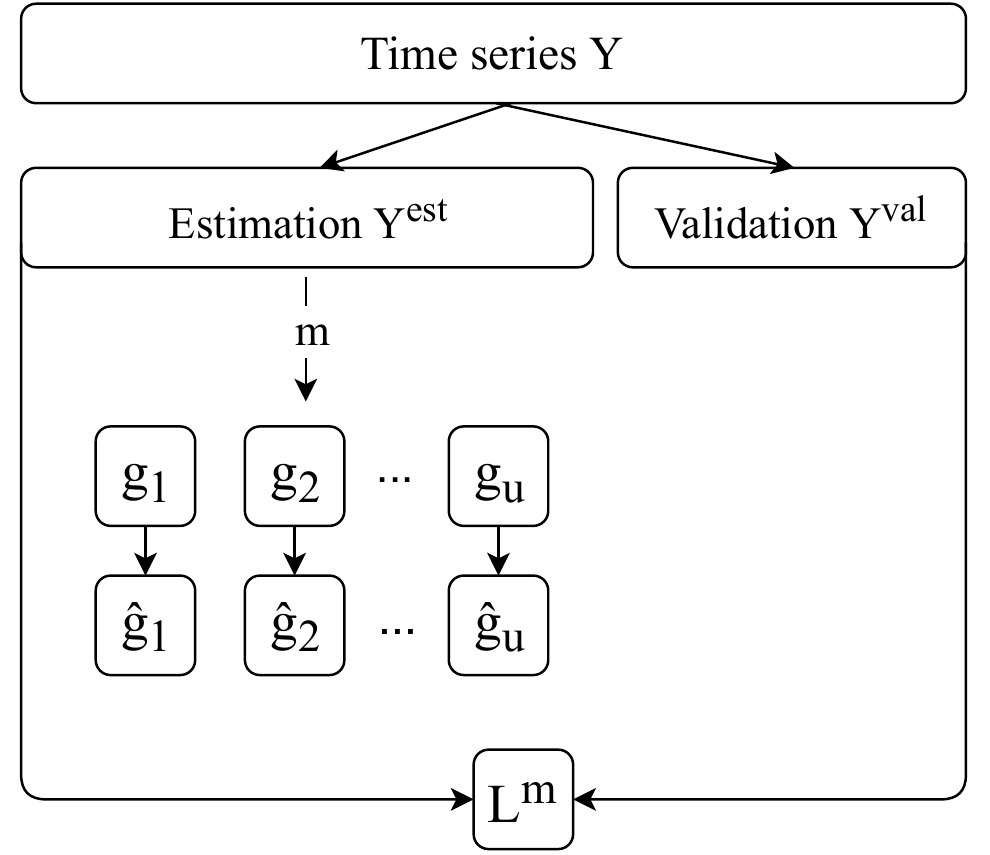}
    \caption{Experimental comparison procedure: A time series is split into an estimation set $Y^{est}$ and a subsequent validation set $Y^{val}$. The first is used to estimate the error $\hat{g}$ that the model $m$ will incur on unseen data, using $z$ different estimation methods. The second is used to compute the actual error $L^m$ incurred by $m$. The objective is to approximate $L^m$ by $\hat{g}$ as well as possible.}
    \label{fig:ts_split}
\end{figure}

The workflow described above is summarised in Figure~\ref{fig:ts_split}. A time series $Y$ is split into an estimation set $Y^{est}$ and a subsequent validation set $Y^{val}$. First, $Y^{est}$ is used to calculate $\hat{g}$, the  estimate of the loss that a predictive model $m$ will incur on future new observations. This is accomplished by further splitting $Y^{est}$ into training and test sets according to the respective estimation procedure $g_i$, $i \in \{1,\dots,z\}$. The model $m$ is built on the training set and $\hat{g}_i$ is computed on the test set.

Second, in order to evaluate the estimates $\hat{g}_i$ produced by the methods $g_i$, $i \in \{1,\dots,z\}$, the model $m$ is re-trained using the complete  set $Y^{est}$ and tested on the validation set $Y^{val}$. Effectively, we obtain $L^m$, the ground truth loss that $m$ incurs on new data.

In summary, the goal of an estimation method $g_i$ is to approximate $L^m$ by $\hat{g}_i$ as well as possible. In Section~\ref{sec:evalmetrics} we describe how to quantify this approximation.

\subsection{Experimental design}

The experimental design was devised to address the following research question: How do the predictive performance estimates of cross-validation methods compare to the estimates of out-of-sample approaches for time series forecasting tasks?

Existing empirical evidence suggests that cross-validation methods provide more accurate estimations than traditionally used OOS approaches in stationary time series forecasting~\cite{bergmeir2012use,bergmeir2014usefulness,bergmeir2015note} (see Section~\ref{sec:lr}). However, many real-world time series comprise complex structures. These include cues from the future that may not have been revealed in the past. Effectively, our hypothesis is that preserving the temporal order of observations when estimating the predictive ability of models is an important component. 

\subsubsection{Embedding dimension and estimation set size}\label{sec:embeddingdim}

We estimate the optimal embedding dimension ($p$) using the method of False Nearest Neighbours~\cite{kennel1992determining}. This method analyses the behaviour of the nearest neighbours as we increase $p$. According to Kennel et al.~\cite{kennel1992determining}, with a low sub-optimal $p$ many of the nearest neighbours will be false. Then, as we increase $p$ and approach an optimal embedding dimension those false neighbours disappear. We set the tolerance of false nearest neighbours to 1\%. The embedding dimension estimated for each series is shown in Table~\ref{tab:data}. Regarding the synthetic case study, we fixed the embedding dimension to 5. The reason for this setup is to try to follow the experimental setup by Bergmeir et al.~\cite{bergmeir2015note}. 

The estimation set ($Y^{est}$) in each time series is the first 70\% observations of the time series -- see Figure~\ref{fig:ts_split}. The validation period is comprised of the subsequent 30\% observations ($Y^{val}$). 

\subsubsection{Estimation methods}

In the experiments we apply a total of 11 performance estimation methods, which are divided into CVAL variants and OOS aproaches. The cross-validation methods are the following:

\begin{description}
    \item[\texttt{CV}] Standard, randomized K-fold cross-validation;
    \item[\texttt{CV-Bl}] Blocked K-fold cross-validation;
    \item[\texttt{CV-Mod}] Modified K-fold cross-validation;
    \item[\texttt{CV-hvBl}] hv-Blocked K-fold cross-validation;
\end{description}

\noindent Conversely, the out-of-sample approaches are the following:

\begin{description}
    \item[\texttt{Holdout}] A simple OOS approach--the first 70\% of $Y^E$ is used for training and the subsequent 30\% is used for testing;
    
    \item[\texttt{Rep-Holdout}] OOS tested in \textit{nreps} testing periods with a Monte Carlo simulation using 70\% of the total observations $t$ of the time series in each test. For each period, a random point is picked from the time series. The previous window comprising 60\% of $t$ is used for training and the following window of 10\% of $t$ is used for testing;
    
    \item[\texttt{Preq-Bls}] Prequential evaluation in blocks in a growing fashion;
    
    \item[\texttt{Preq-Sld-Bls}] Prequential evaluation in blocks in a sliding fashion--the oldest block of data is discarded after each iteration;
    
    \item[\texttt{Preq-Bls-Gap}] Prequential evaluation in blocks in a growing fashion with a gap block--this is similar to the method above, but comprises a block separating the training and testing blocks in order to increase the independence between the two parts of the data;
    
    \item[\texttt{Preq-Grow} and \texttt{Preq-Slide}] As baselines we also include the exhaustive prequential methods in which an observation is first used to test the predictive model and then to train it. We use both a growing/landmark window (\texttt{Preq-Grow}) and a sliding window (\texttt{Preq-Slide}). 
\end{description}

We refer to Section~\ref{sec:lr} for a complete description of the methods. The number of folds $K$ or repetitions $nreps$ in these methods is 10, which is a commonly used setting in the literature. The number of observations removed in \texttt{CV-Mod} and \texttt{CV-hvBl} (c.f. Section~\ref{sec:lr}) is the embedding dimension $p$ in each time series.

\subsubsection{Evaluation metrics}\label{sec:evalmetrics}

Our goal is to study which estimation method provides a $\hat{g}$ that best approximates $L^m$. Let $\hat{g}^m_i$ denote the estimated loss by the learning model $m$ using the estimation method $g$ on the estimation set, and $L^m$ denote the ground truth loss of learning model $m$ on the test set. The objective is to analyze how well $\hat{g}^m_i$ approximates $L^m$. This is quantified by the absolute predictive accuracy error (APAE) metric and the predictive accuracy error (PAE)~\cite{bergmeir2015note}:

\begin{equation}\label{eq:apae}
    \text{APAE} = |\hat{g}^m_i - L^m|
\end{equation}

\begin{equation}\label{eq:pae}
    \text{PAE} = \hat{g}^m_i - L^m 
\end{equation}

The APAE metric evaluates the error size of a given estimation method. On the other hand, PAE measures the error bias, i.e., whether a given estimation method is under-estimating or over-estimating the true error.

Another question regarding evaluation is how a given learning model is evaluated regarding its forecasting accuracy. In this work we evaluate models according to root mean squared error (RMSE). This metric is traditionally used for measuring the differences between the estimated values and actual values. 

\subsubsection{Learning algorithm}

The results shown in this work are obtained using a rule-base regression system Cubist~\cite{Cubist2014}, a variant of Quinlan's model tree~\cite{quinlan1993combining}. This method presented the best forecasting results among several other predictive models in a recent study~\cite{cerqueira2018arbitrage}. Notwithstanding, other learning algorithms were tested, namely the lasso~\cite{tibshirani1996regression} and a random forest~\cite{ranger2015}. The conclusions drawn using these algorithms are similar to the ones reported in the next sections.

\section{Empirical experiments}\label{sec:ee}

\subsection{Results with synthetic case study}\label{sec:STS_results}

\begin{figure}[t]
    \centering
    \includegraphics[width=.8\textwidth, trim=0cm 0cm 0cm 0cm, clip=TRUE]{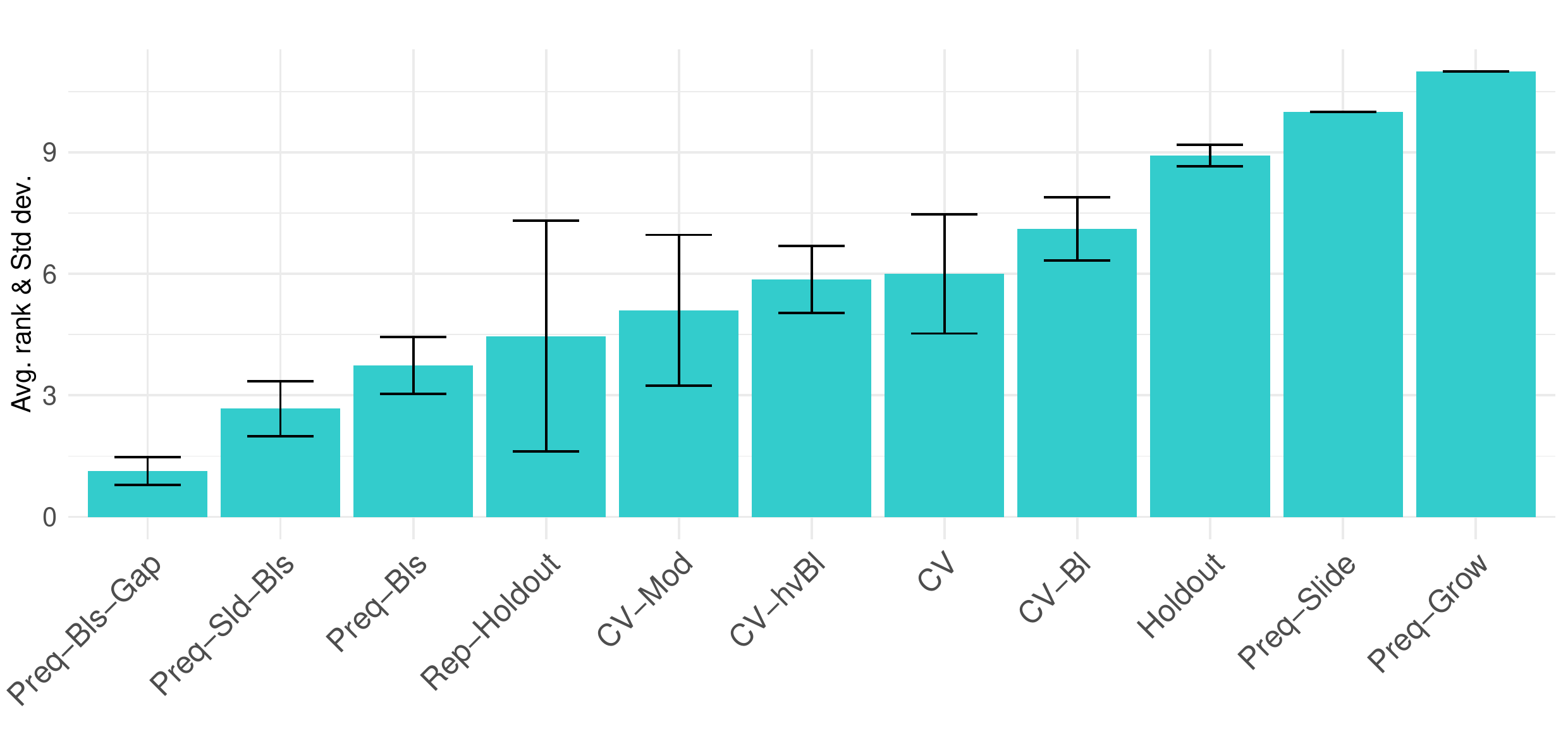}
    \caption{Average rank and respective standard deviation of each estimation methods in case study S1}
    \label{fig:s1avg}
\end{figure}

\begin{figure}[bh]
    \centering
    \includegraphics[width=.8\textwidth, trim=0cm 0cm 0cm 0cm, clip=TRUE]{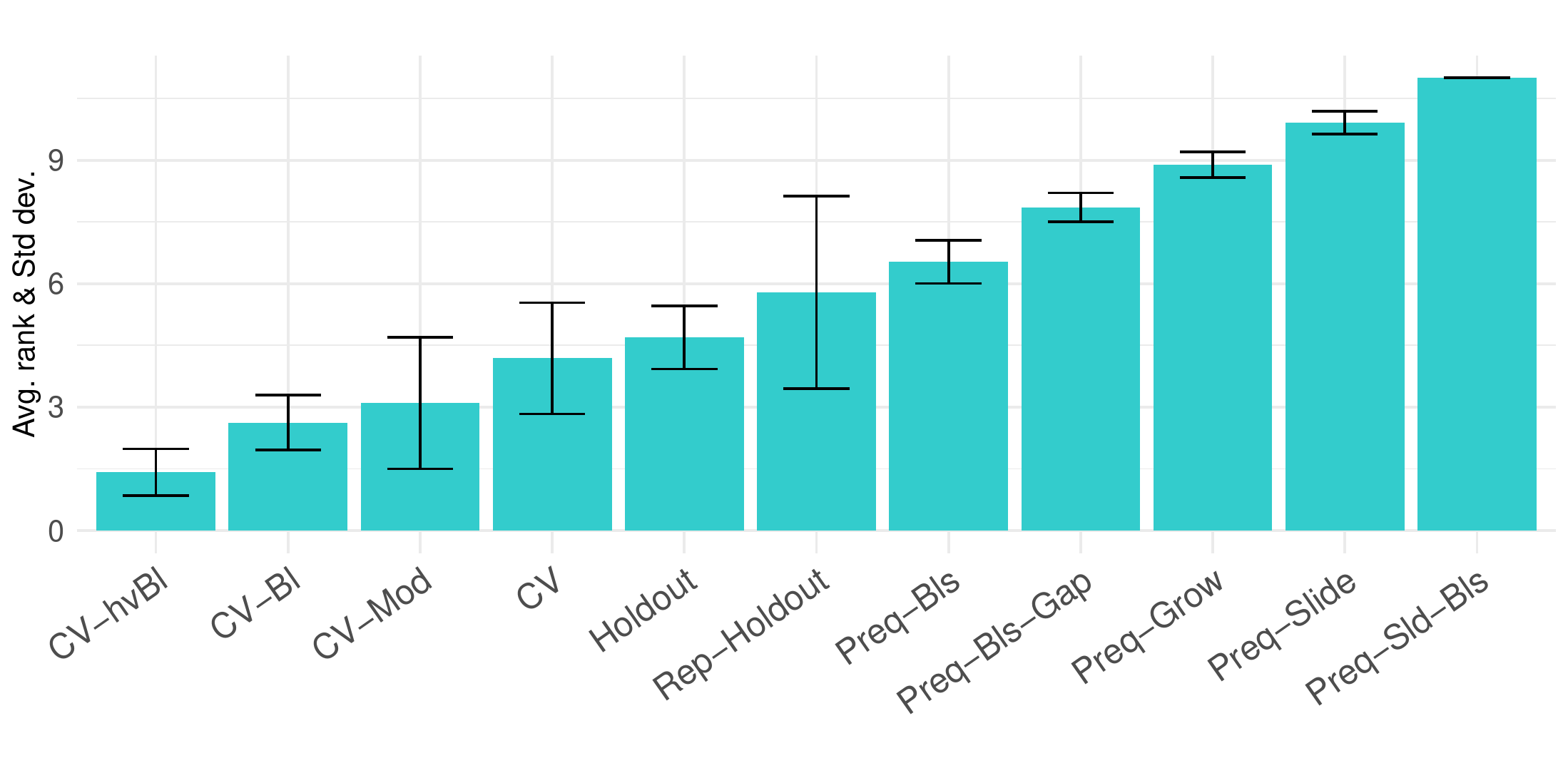}
    \caption{Average rank and respective standard deviation of each estimation methods in case study S2}
    \label{fig:s2avg}
\end{figure}

\begin{figure}[hbt]
    \centering
    \includegraphics[width=.8\textwidth, trim=0cm 0cm 0cm 0cm, clip=TRUE]{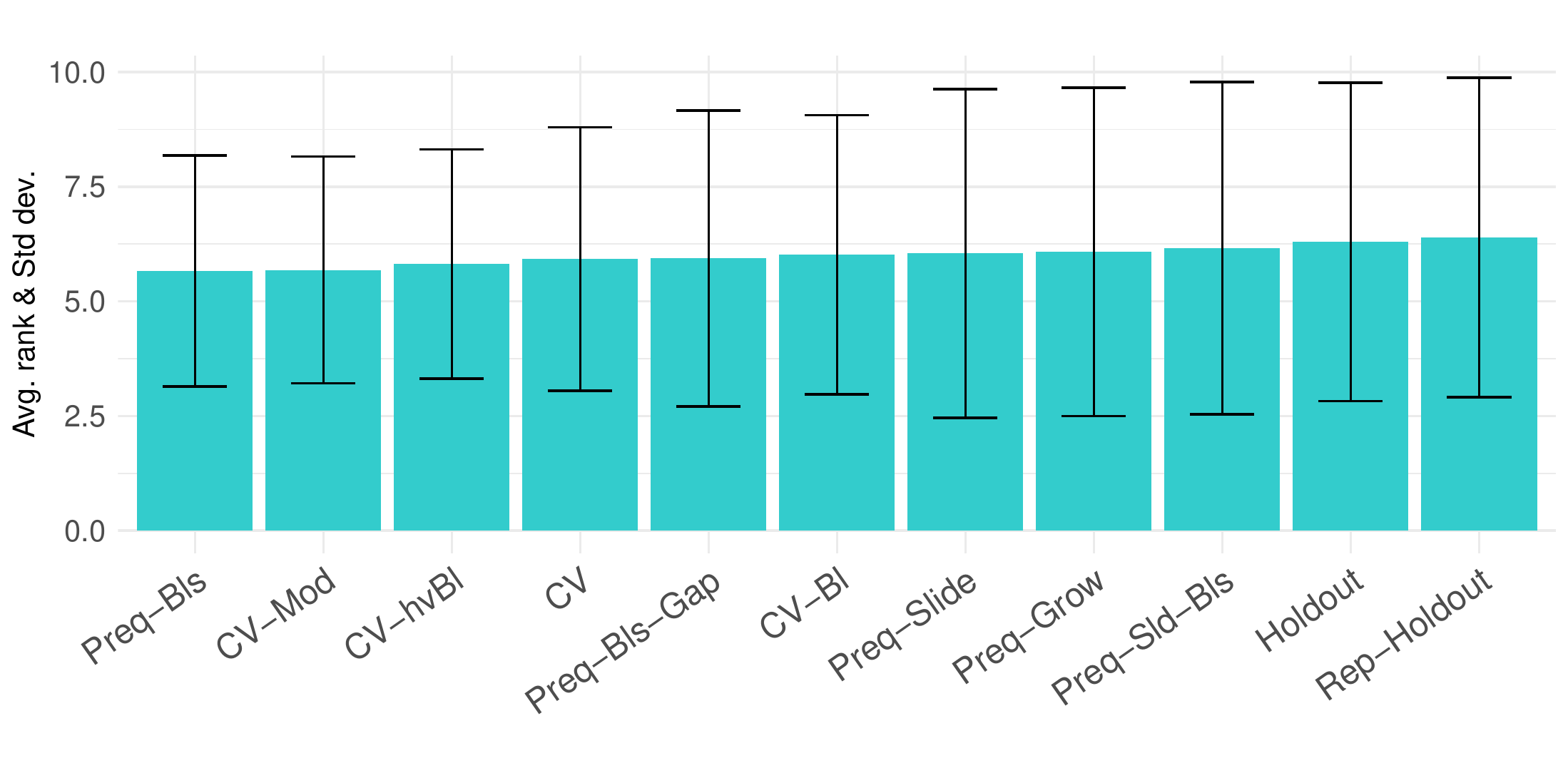}
    \caption{Average rank and respective standard deviation of each estimation methods in case study S3}
    \label{fig:s3avg}
\end{figure}

In this section we start by analysing the average rank, and respective standard deviation, of each estimation method and for each synthetic scenario (S1, S2, and S3), according to the metric APAE. For example, a rank of 1 in a given Monte Carlo repetition means that the respective method was the best estimator in that repetition. These analyses are reported in Figures~\ref{fig:s1avg}--\ref{fig:s3avg}. This initial experiment is devised to reproduce the results by Bergmeir et al.~\cite{bergmeir2015note}. Later, we will analyse how these results compare when using real-world time series.

The results shown by the average ranks corroborate those presented by Bergmeir et al.~\cite{bergmeir2015note}. That is, cross validation approaches generally perform better (i.e., show a lower average rank) relative to the simple out-of-sample procedure \texttt{Holdout}. This can be concluded from all three scenarios: S1, S2, and S3.

Focusing on scenario S1, the estimation method with the best average rank is \texttt{Preq-Bls-Gap}, followed by the other two prequential variants (\texttt{Preq-Sld-Bls}, and \texttt{Preq-Bls}). Although the \texttt{Holdout} procedure is clearly a relatively poor estimator (worst average rank), the repeated holdout in multiple testing periods (\texttt{Rep-Holdout}) shows a better average rank than the cross validation procedures (though with a large standard deviation). Among cross validation procedures, \texttt{CV-Mod} presents the best average rank.

Scenario S2 shows a seemingly different story relative to S1. In this problem, the prequential variants present the worst average rank, while the cross validation procedures show the best estimation ability. Among all, \texttt{CV-hvBl} shows the best average rank. Moreover, \texttt{Rep-Holdout} presents again a large standard deviation in rank, relative to the remaining estimation methods.

Regarding the scenario S3, the outcome is less clear than the previous two scenarios. The methods show a closer average rank among them, with large standard deviations.

In summary, this first experiment corroborates the experiment carried our by Bergmeir et al.~\cite{bergmeir2015note}. Notwithstanding, other methods that the authors did not test show an interesting estimation ability in these particular scenarios, namely the prequential variants.

The synthetic scenarios comprise time series that are stationary. However, real-world time series often comprise complex dynamics that break stationarity. When choosing a performance estimation method one should take this issue into consideration. To account for time series stationarity, in the next section we analyze the estimation methods using real-world time series. We will also control for time series stationarity to study its impact on the results.

\subsection{Results with real-world case study}\label{sec:RWTS_results}

In this section we analyze the performance estimation ability of each method using a case study comprised of real-world time series from different domains.

\begin{figure}[ht]
    \centering
    \includegraphics[width=.8\textwidth, trim=0cm 0cm 0cm 0cm, clip=TRUE]{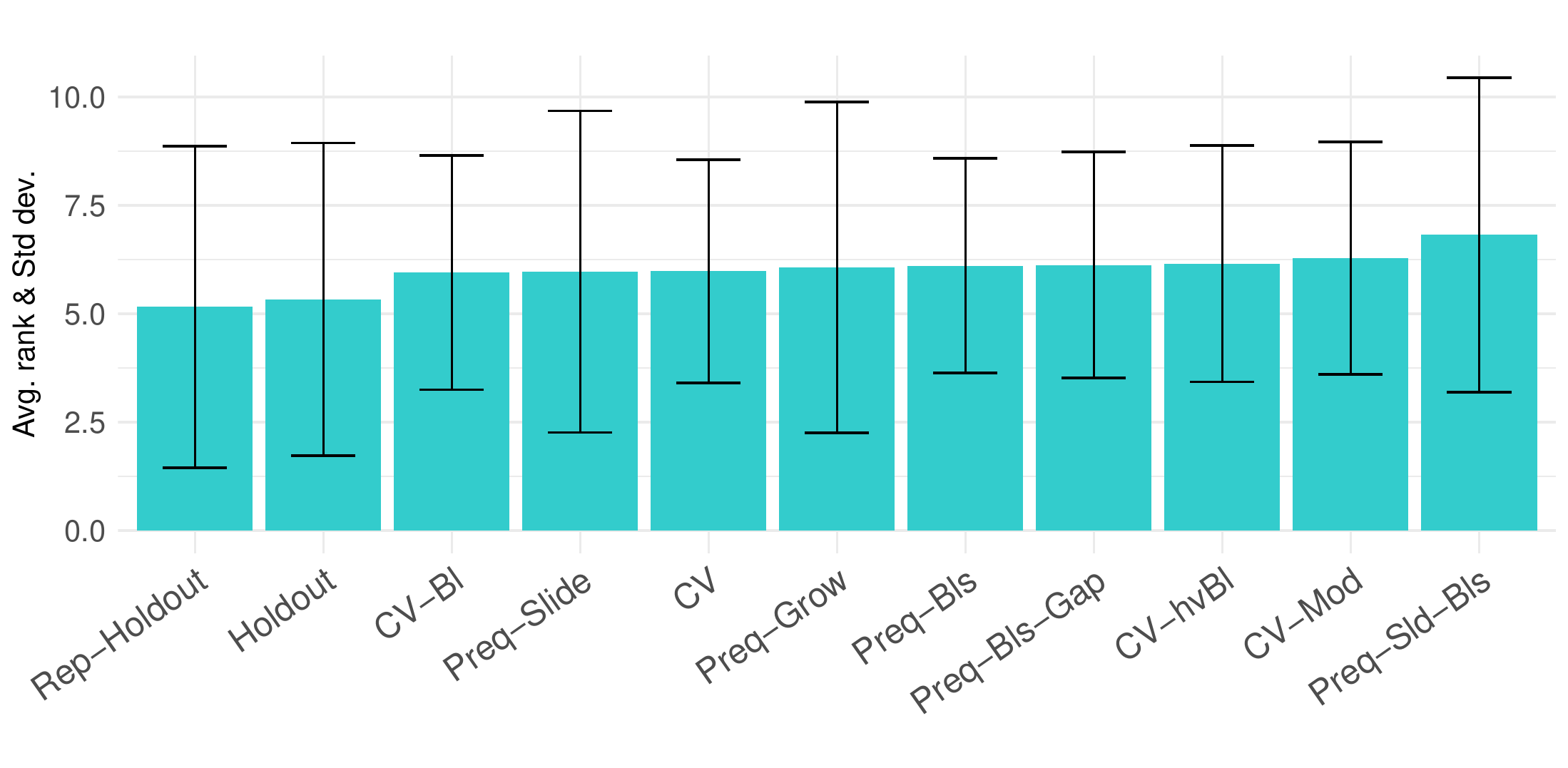}
    \caption{Average rank and respective standard deviation of each estimation methods in case study RWTS}
    \label{fig:rw_avg}
\end{figure}

\subsubsection{Main results}

To accomplish this in Figure~\ref{fig:rw_avg} we start by analyzing the average rank, and respective standard deviation, of each estimation method using the APAE metric. This graphic tells a different story relative to the synthetic case study. Particularly, the \texttt{Rep-Holdout} and \texttt{Holdout} show the best estimation ability in terms of the average rank. The method \texttt{CV-Bl} is the best estimator among the cross-validation procedures.

\begin{figure}[!h]
    \centering
    \includegraphics[width=.8\textwidth, trim=0cm 0cm 0cm 0cm, clip=TRUE]{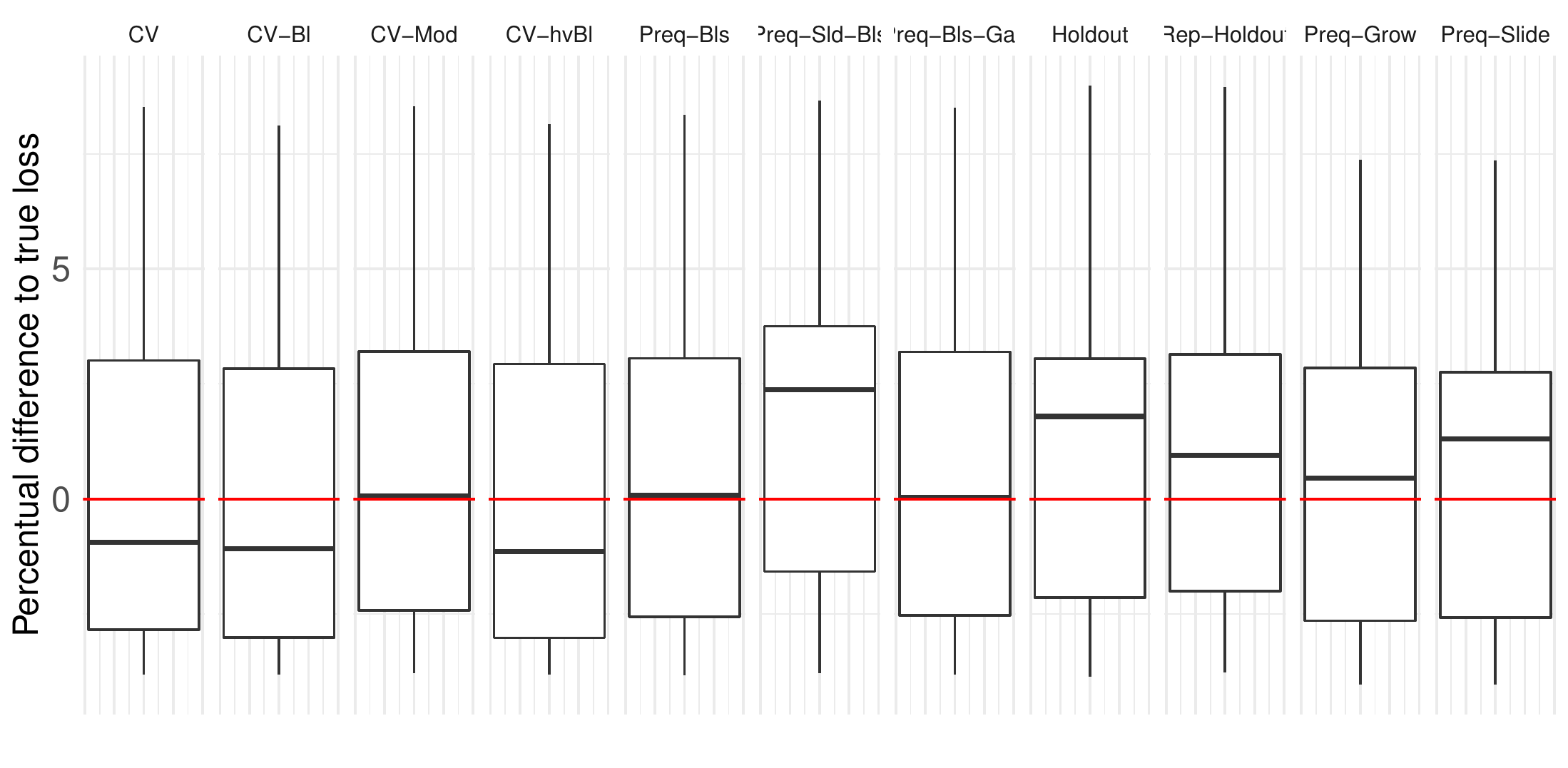}
    \caption{Percentual difference of the estimated loss relative to the true loss for each estimation method in the RWTS case study. Values below the zero line represent under-estimations of error. Conversely, values above the zero line represent over-estimations of error.}
    \label{fig:rwpd}
\end{figure}

\begin{figure}[tbh]
    \centering
    \includegraphics[width=.8\textwidth, trim=0cm 0cm 0cm 0cm, clip=TRUE]{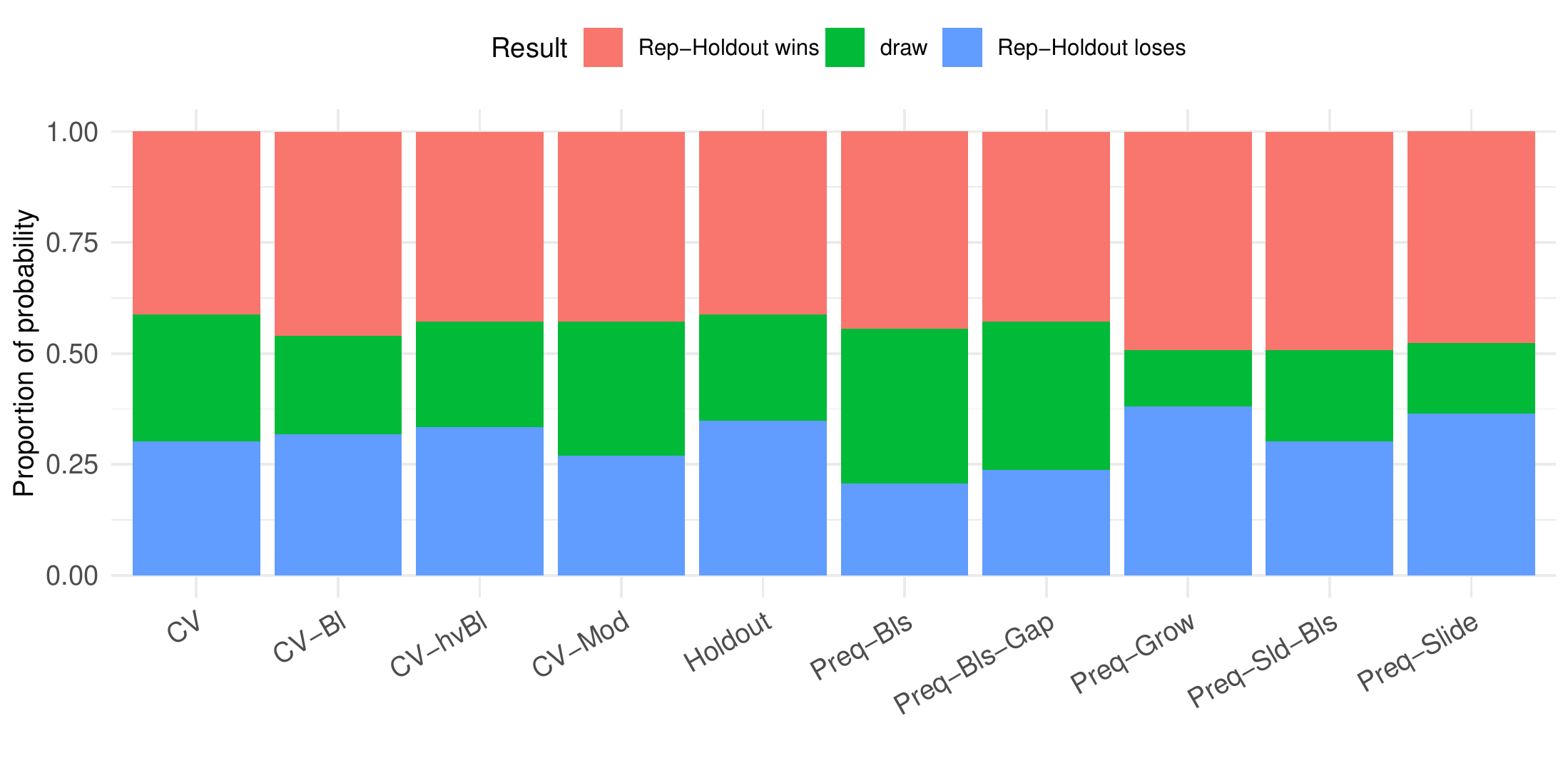}
    \caption{Proportion of probability of the outcome when comparing the performance estimation ability of the respective estimation method with the \texttt{Rep-Holdout} method. The probabilities are computed using the Bayes sign test.}
    \label{fig:propall}
\end{figure}

In order to study the direction of the estimation error, in Figure~\ref{fig:rwpd} we present for each method the percentual difference between the estimation error and the true error according to the PAE metric. In this graphic, values below the zero line denote under-estimations of error, while values above the zero line represent over-estimations. In general, cross-validation procedures tend to under-estimate the error (i.e. are optimistic estimators), while the prequential and out-of-sample variants tend to over-estimate the error (i.e. are pessimistic estimators).

This result corroborates the results on Twitter time-ordered data \cite{mozetic2018evaluate}. The authors found that all variants of cross-validation procedures tend to under-estimate the errors, while the out-of-sample procedures tend to over-estimate them.

We also study the statistical significance of the obtained results  in terms of error size (APAE) according to a Bayesian analysis~\cite{benavoli2017time}. Particularly, we employed the Bayes sign test to compare pairs of methods across multiple problems. We define the \textit{region of practical equivalence}~\cite{benavoli2017time} (ROPE) to be the interval [-2.5\%, 2.5\%] in terms of APAE. Essentially, this means that two methods show indistinguishable performance if the difference in performance between them falls within this interval. For a thorough description of the Bayesian analysis for comparing predictive models we refer to the work by Benavoli et al~\cite{benavoli2017time}.

In this experiment we fix the method \texttt{Rep-Holdout} as the baseline, since it is the one showing the best average rank (Figure~\ref{fig:rw_avg}). According to the illustration in Figure~\ref{fig:propall}, the probability of \texttt{Rep-Holdout} winning (i.e., showing a significantly better estimation ability) is generally larger than the opposite.

\subsubsection{Controlling for stationarity}

After analyzing the synthetic case study we hypothesized that the results were biased due to the stationarity assumption. In this section we repeat the average rank experiment in the real-world case study controlling for stationarity. We consider a time series stationary according to the analysis carried out in Section~\ref{sec:stationarity}.

\begin{figure}[tbh]
    \centering
    \includegraphics[width=.8\textwidth, trim=0cm 0cm 0cm 0cm, clip=TRUE]{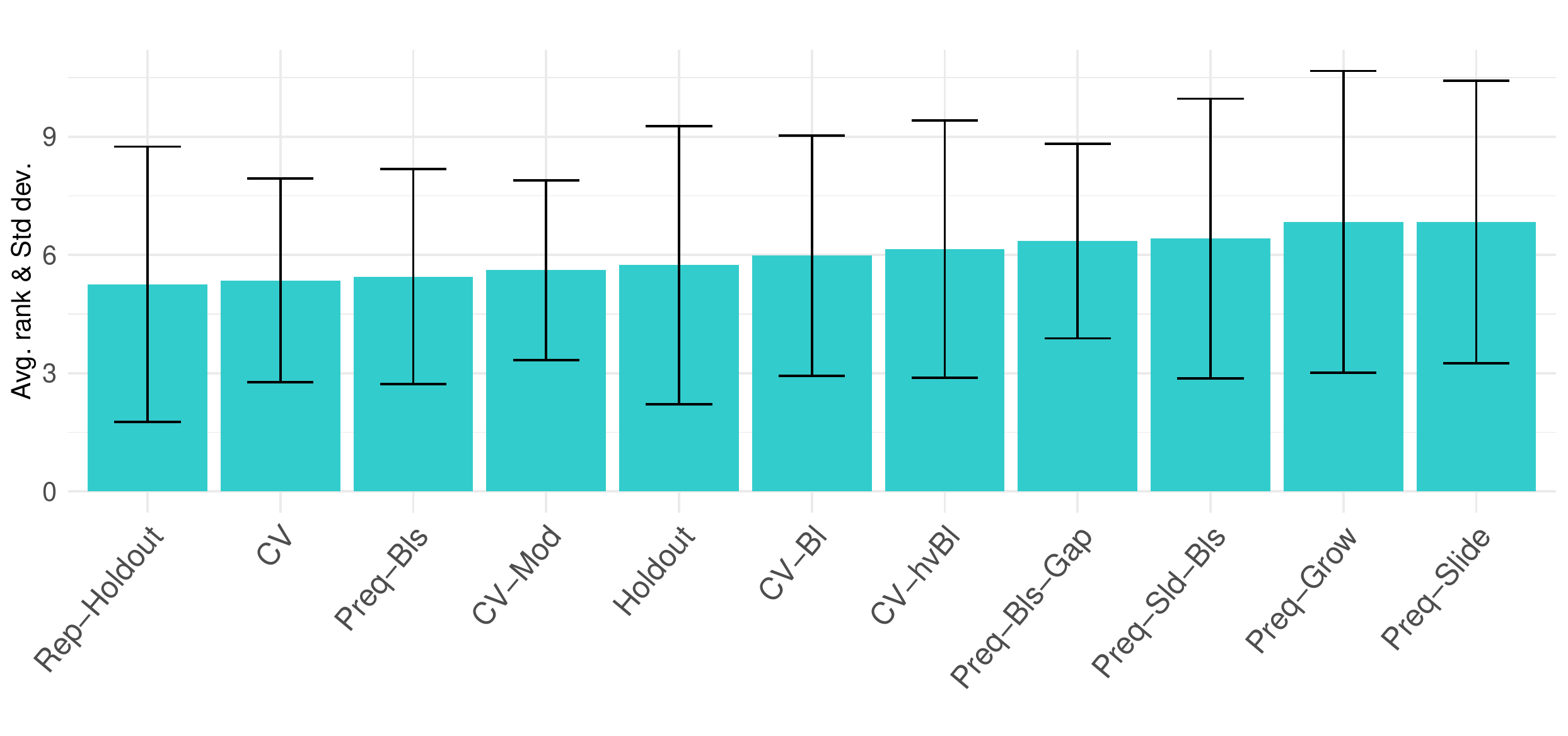}
    \caption{Average rank and respective standard deviation of each estimation methods in case study RWTS for stationary time series (31 time series).}
    \label{fig:rw_stat_avg}
\end{figure}

In Figure~\ref{fig:rw_stat_avg} we present the results considering only the real world time series that are stationary. According to the average rank, the typical cross-validation approach \texttt{CV} presents the best estimation ability,  followed by \texttt{Rep-Holdout}.

\begin{figure}[tbh]
    \centering
    \includegraphics[width=.8\textwidth, trim=0cm 0cm 0cm 0cm, clip=TRUE]{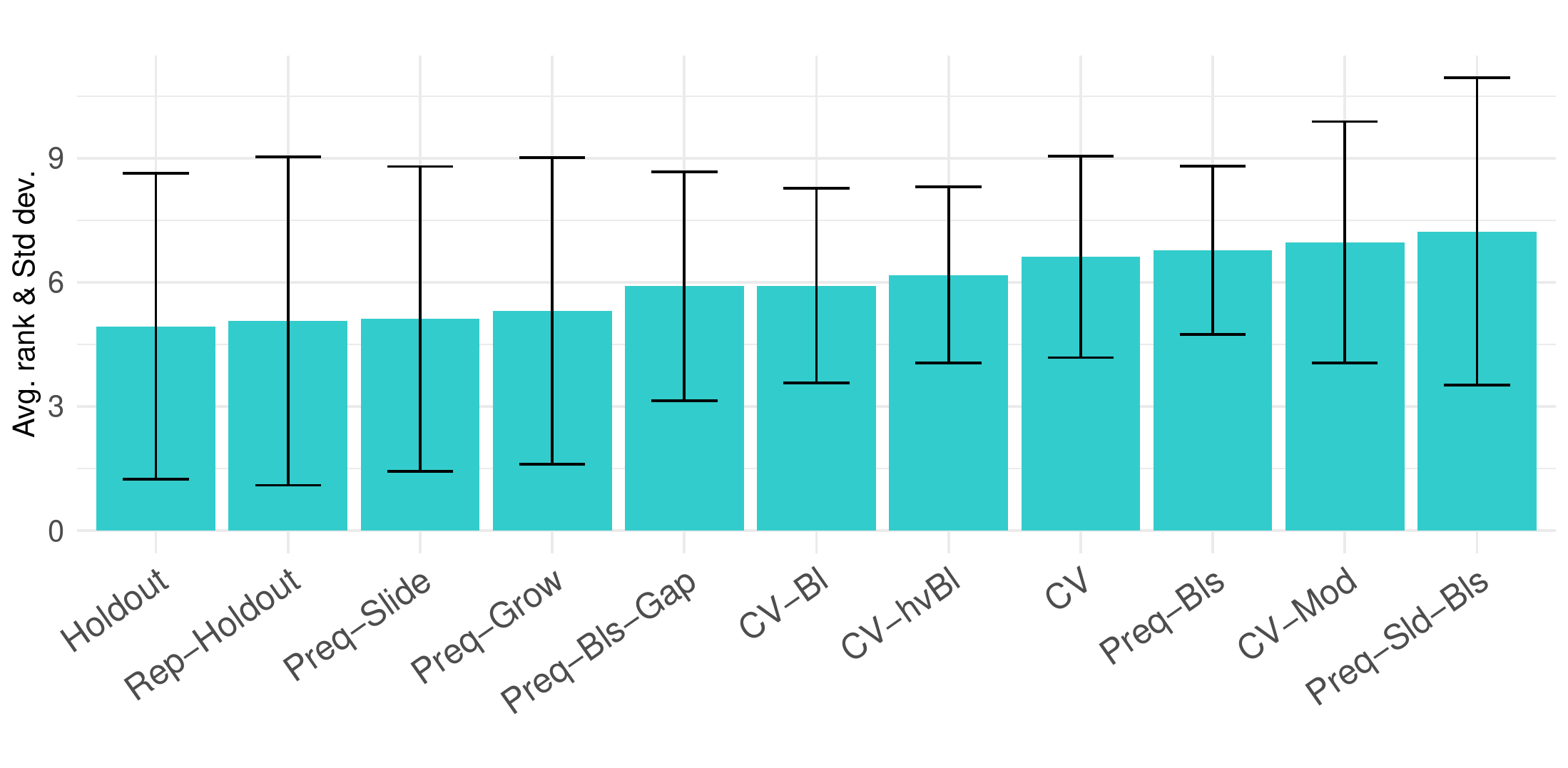}
    \caption{Average rank and respective standard deviation of each estimation methods in case study RWTS for non-stationary time series (31 time series).}
    \label{fig:rw_nonstat_avg}
\end{figure}

In Figure~\ref{fig:rw_nonstat_avg} we present a similar analysis for the non-stationary time series, whose results are considerably different relative to stationary time series. In this scenario, \texttt{CV} is one of the worst estimator according to average rank. The out-of-sample approaches \texttt{Holdout} and \texttt{Rep-Holdout} present the best estimation ability.

\subsubsection{Descriptive model}

What makes an estimation method appropriate for a given time series is related to the characteristics of the data. For example, in the previous section we analyzed the impact that stationarity has in terms of what is the best estimation method.

The real-world time series case study comprises a set of time series from different domains. In this section we present, as a descriptive analysis, a tree-based model that relates some characteristics of time series according with the most appropriate estimation method for that time series. Basically, we create a predictive task in which the attributes are some characteristics of a time series, and the categorical target variable is the estimation method that best approximates the true loss in that time series. We use CART~\cite{breiman2017classification} (classification and regression tree) algorithm for obtaining the  model for this task. The characteristics used as predictor variables are the following summary statistics: 

\begin{itemize}
	\item \textbf{Skewness}, for measuring the symmetry of the distribution of the time series;
	\item 5-th and 95-th Percentiles (\textbf{Perc05}, \textbf{Perc95}) of the standardized time series;
	\item Acceleration (\textbf{Accel}.), as the average ratio between a simple moving average and the exponential moving average;
	\item Inter-quartile range (\textbf{IQR}), as a measure of the spread of the standardized time series;
	\item Serial correlation, estimated using a Box-Pierce test statistic;
	\item Long-range dependence, using a Hurst exponent estimation with wavelet transform;
	\item Maximum Lyapunov Exponent, as a measure of the level of chaos in the time series;
	\item a boolean variable, indicating whether or not the respective time series is stationary according to the wavelet spectrum test~\cite{nason2013test}.
\end{itemize}

\begin{figure}[ht]
    \centering
    \includegraphics[width=\textwidth, trim=0cm .5cm 0cm 0cm, clip=TRUE]{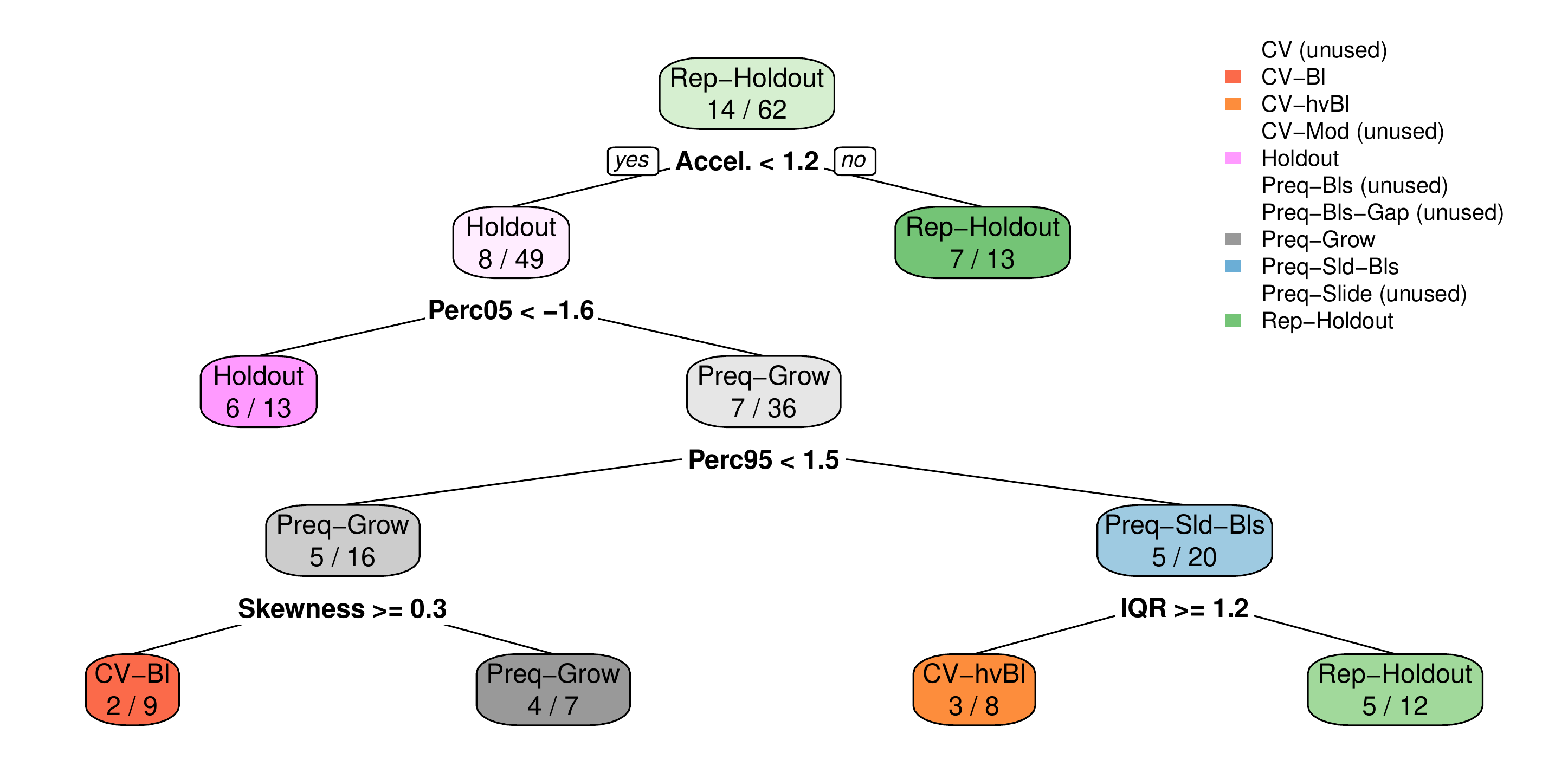}
    \caption{Decision tree that maps the characteristics of time series to the most appropriate estimation method. Graphic created using the \textit{rpart.plot} framework~\cite{rpartplot}.}
    \label{fig:descriptivemodel}
\end{figure}

\noindent The characteristics used in  the obtained decision tree are written in boldface. The decision tree is shown in Figure \ref{fig:descriptivemodel}. The numbers below the name of the method in each node denote the number of times the respective method is best over the number of time series covered in that node.


Some of the estimation methods do not appear in the tree model. The tree leaves, which represent a decision, are dominated by the \texttt{Rep-Holdout} and \texttt{Holdout} estimation methods. The estimation methods \texttt{CV-Bl}, \texttt{Preq-Slide}, \texttt{Preq-Grow}, and \texttt{CV-hvBl} also appear in other leaves.

The estimation method in the root node is \texttt{Rep-Holdout}, which is the best method most of the times across the 62 time series. The first split is performed according to the acceleration characteristic of time series. Basically, if acceleration is not below 1.2, the tree leads to a leaf node with \texttt{Rep-Holdout} as the most appropriate estimation method. Otherwise, the tree continues with more tests in order to find the most suitable estimation method for each particular scenario.

\section{Discussion}\label{sec:disc}

\subsection{Impact of the results}

In the experimental evaluation we compare several performance estimation methods in two distinct scenarios: (1) a synthetic case study in which artificial data generating processes are used to create stationary time series; and (2) a real-world case study comprising 62 time series from different domains. The synthetic case study is based on the experimental setup used in previous studies by Bergmeir et al. for the same purpose of evaluating performance estimation methods for time series forecasting tasks~\cite{bergmeir2012use,bergmeir2014usefulness,bergmeir2015note}.

Bergmeir et al. show in previous studies~\cite{bergmeir2011forecaster,bergmeir2012use} that the blocked form of cross-validation, denoted here as \texttt{CV-Bl}, yields more accurate estimates than a simple out-of-sample evaluation (\texttt{Holdout}) for stationary time series forecasting tasks. The method \texttt{CV} is also suggested to be ``a better choice than OOS[\texttt{Holdout}] evaluation" as long as the data are well fitted by the model~\cite{bergmeir2015note}. To some extent part of the results from our experiments corroborate these conclusions. Specifically, this is verified by the APAE incurred by the estimation procedures in the synthetic case studies. 

However, according to our experiments, the results from the synthetic stationary case studies do not reflect those obtained using real-world time series. In general, holdout applied with multiple randomized testing periods (\texttt{Rep-Holdout}) provides the most accurate performance estimates. Notwithstanding, for stationary time series \texttt{CV} also shows a competitive estimation ability.

In a real-world environment we are prone to deal with time series with complex structures and different sources of non-stationary variations. These comprise nuances of the future that may not have revealed themselves in the past~\cite{tashman2000out}. Consequently, we believe that in these scenarios, \texttt{Rep-Holdout} is a better option as performance estimation method relative to cross-validation approaches.

\subsection{On the importance of data size}

The temporal order preservation by OOS approaches, albeit more realistic, comes at a cost since less data is available for estimating predictive performance. As Bergmeir et al.~\cite{bergmeir2015note} argue, this may be important for small data sets, where a more efficient use of the data (e.g. \texttt{CV}) may be beneficial. However, during our experimental evaluation we did not found compelling evidence to back this claim. In the reported experiments we fixed the data size to 200 observations, as Bergmeir et al~\cite{bergmeir2015note} did. In order to control for data size, we varied this parameter from a size of 100 to a size of 3000, by intervals of 100 (100, 200, ..., 3000). The experiments did not provide any evidence that the size of the synthetic time series had a noticeable effect on the error of estimation methods.

In our experiments the size of the time series in the real-world case study are in the order of a few thousands. For large scale data sets the recommendation by Dietterich~\cite{dietterich1998approximate}, and usually adopted in practice, is to apply a simple out-of-sample estimation procedure (\texttt{Holdout}).

\subsection{Scope of the real-world case study}

In this work we center our study on univariate numeric time series. Nevertheless, we believe that the conclusions of our study are independent of this assumption and should extend for other types of time series. The objective is to predict the next value of the time series, assuming immediate feedback from the environment. Moreover, we focus on time series with a high sampling frequency, specifically, half-hourly, hourly, and daily data. The main reason for this is because high sampling frequency is typically associated with more data, which is important for fitting the predictive models from a machine learning point of view. Standard forecasting benchmark data are typically more centered around low sampling frequency time series, for example the M competition data~\cite{makridakis1982accuracy}.

\section{Final remarks}\label{sec:fr}

In this paper we analyse the ability of different methods to approximate the loss that a given predictive model will incur on unseen data. This error estimation process is performed in every machine learning task for model selection and hyper-parameter tuning. We focus on performance estimation for time series forecasting tasks. Since there is currently no settled approach for performance estimation in these settings, our objective is to compare different available methods and test their suitability. 

We analyse several methods that can be generally split into out-of-sample approaches and cross-validation methods. These were applied to two case studies: a synthetic environment with stationary time series and a real-world scenario with potential non-stationarities.

In a stationary setting the cross-validation variants are shown to have a competitive estimation ability. However, when non-stationarities are present, they systematically provide worse estimations than the out-of-sample approaches.

Bergmeir et al.~\cite{bergmeir2012use,bergmeir2014usefulness,bergmeir2015note} suggest that for stationary time series one should use cross-validation in a blocked form (\texttt{CV-Bl}). On the other hand, for real-world time series with potential non-stationarities we conclude that approaches that maintain the temporal order of data provide better error estimations. In particular, out-of-sample applied in multiple testing periods (\texttt{Rep-Holdout}) is recommended. In the interest of reproducibility, the methods and data sets are publicly available at \url{https://github.com/vcerqueira/performance_estimation}.


\bibliographystyle{spmpsci}      
\bibliography{performance_estimation_tsf}   

\section*{Appendix}\label{appendix}

\begin{table}[!thb]
	\centering
	\caption{Time series data sets and respective summary. The column p denotes the size of the embedding dimension, I denotes the number of differences applied to the time series to make it trend-stationary, and S represents whether or not the de-trended time series is stationary (1 if it is).}		
	\resizebox{\textwidth}{!}{%
	\begin{tabular}{cp{4.5cm}p{3.5cm}p{6cm}llll}
	\toprule
	\textbf{ID} & \textbf{Time series} & \textbf{Data source} & \textbf{Data characteristics} & \textbf{Size} & \textbf{p} & \textbf{I} & \textbf{S}\\ 
	\midrule
	    1 & Rotunda AEP & \multirow{3}{4.0cm}{\strut Porto Water Consumption from different locations in the city of Porto~\cite{cerqueira2017ecml} } &  \multirow{3}{*}{\parbox{5.8cm}{Half-hourly values from Nov. 11, 2015 to Jan. 11, 2016}} & 3000 & 30 & 0 & 0\\
	    2 & Preciosa Mar & & & 3000 & 9 & 1 & 0\\
	    3 & Amial & & & 3000 & 11 & 0 & 0\\
	    
		\midrule 
		
		4 & Global Horizontal Radiation & \multirow{4}{4.0cm}{Solar Radiation Monitoring~\cite{cerqueira2017ecml}} &  \multirow{4}{*}{\parbox{5.8cm}{Hourly values from Apr. 25, 2016 to Aug. 25, 2016}} & 3000 & 23 & 1 & 0\\
		5 & Direct Normal Radiation & & & 3000 & 19 & 1 & 1\\
		6 & Diffuse Horizontal Radiation & & & 3000 & 18 & 1 & 1\\
		7 & Average Wind Speed & & & 3000 & 10 & 1 & 0\\
            \midrule
		8 & Humidity & \multirow{3}{4.0cm}{Bike Sharing~\cite{cerqueira2017ecml}} & & 1338 & 11 & 0 & 0\\
		9 & Windspeed &  & Hourly values from Jan. 1, 2011 & 1338 & 12 & 0 & 1\\ 
		10 & Total bike rentals & & Mar. 01, 2011 & 1338 & 8 & 0 & 1\\
		\midrule    
		11 & AeroStock 1 & \multirow{10}{4.0cm}{\strut Stock price values from different aerospace companies \cite{cerqueira2017ecml}} &  \multirow{10}{*}{\parbox{5.8cm}{\strut Daily stock prices from January 1988 through October 1991}} & 949 & 6 & 1 & 1\\
		12 & AeroStock 2 & & & 949 & 13 & 1 & 0\\
		13 & AeroStock 3 & & & 949 & 7 & 1 & 1\\
		14 & AeroStock 4 & & & 949 & 8 & 1 & 1\\
		15 & AeroStock 5 & & & 949 & 6 & 1 & 1\\
		16 & AeroStock 6 & & & 949 & 10 & 1 & 1\\
		17 & AeroStock 7 & & & 949 & 8 & 1 & 1\\
		18 & AeroStock 8 & & & 949 & 8 & 1 & 1\\
		19 & AeroStock 9 & & & 949 & 9 & 1 & 1\\
		20 & AeroStock 10 & & & 949 & 8  & 1 & 1\\
		\midrule
		21 & CO.GT & \multirow{13}{4.0cm}{\strut Air quality indicators in an Italian city \cite{Lichman:2013}} &  \multirow{13}{*}{\parbox{5.8cm}{\strut Hourly values from Mar. 10, 2004 to Apr. 04 2005}} & 3000 & 30  & 1 & 0\\
		22 & PT08.S1.CO & & & 3000 & 8 & 1 & 0\\
		23 & NMHC.GT & & & 3000 & 10 & 1 & 0\\
		24 & C6H6.GT & &  & 3000 & 13 & 0 & 1\\
		25 & PT08.S2.NMHC & &  & 3000 & 9 & 0 & 0\\
		26 & NOx.GT & & & 3000 & 10  & 1 & 1\\
		27 & PT08.S3.NOx & &  & 3000 & 10 & 1 & 0\\
		28 & NO2.GT & &  & 3000 & 30 & 1 & 0\\
		29 & PT08.S4.NO2 & &  & 3000 & 8 & 0 & 0\\
		30 & PT08.S5.O3 & &  & 3000 & 8 & 0 & 1\\
		31 & Temperature & &  & 3000 & 8 & 1 & 0\\
		32 & RH & & & 3000 & 23 & 1 & 0\\
		33 & Humidity & & & 3000 & 10  & 1 & 0\\

		\bottomrule    
	\end{tabular}%
	}
	\label{tab:data}
\end{table}

\begin{table}[!thb]
	\centering
	\caption{Continuation of Table 1}		
	\resizebox{\textwidth}{!}{%
	\begin{tabular}{cp{4.5cm}p{3.5cm}p{6cm}llll}
	\toprule
	\textbf{ID} & \textbf{Time series} & \textbf{Data source} & \textbf{Data characteristics} & \textbf{Size} & \textbf{p} & \textbf{I} & \textbf{S}\\ 
	\midrule    
	    34 & Electricity Total Load & \multirow{5}{4.0cm}{Hospital Energy Loads~\cite{cerqueira2017ecml}} &  \multirow{5}{*}{\parbox{5.8cm}{Hourly values from Jan. 1, 2016 to Mar. 25, 2016}} & 3000 & 19 & 0 & 1\\
	    35 & Equipment Load & & & 3000 & 30 & 0 & 1\\
	    36 & Gas Energy & & & 3000 & 10 & 1 & 1\\
	    37 & Gas Heat Energy & & & 3000 & 13 & 1 & 1\\
	    38 & Water heater Energy & & & 3000 & 30 & 0 & 1\\
		39 & Total Demand & \multirow{2}{4.0cm}{Australian Electricity \cite{koprinska2011yearly}} &  \multirow{2}{*}{\parbox{5.8cm}{Half-hourly values from Jan. 1, 1999 to Mar. 1, 1999}} & 2833 & 6 & 0 & 1\\
		40 & Recommended Retail Price & & & 2833 & 19 & 0 & 0\\		 
		\midrule
		41 & Sea Level Pressure & \multirow{3}{4.0cm}{Ozone Level Detection \cite{Lichman:2013}} &  \multirow{3}{*}{\parbox{5.8cm}{Daily values from Jan. 2, 1998 to Dec. 31, 2004}} & 2534 & 9 & 0 & 1\\
		42 & Geo-potential height & & & 2534 & 7 & 0 & 1\\
		43 & K Index & & & 2534 & 7 & 0 & 1\\
		44 & Flow of Vatnsdalsa river & \multirow{17}{4.0cm}{Data market~\cite{datamarket}} & Daily, from Jan. 1, 1972 to Dec. 31, 1974 & 1095 & 11 & 0 & 0\\
		45 & Rainfall in Melbourne &  & Daily, from from 1981 to 1990 & 3000 & 29 & 0 & 0\\
		
		46 & Foreign exchange rates &  & Daily, from Dec. 31, 1979 to Dec. 31, 1998 & 3000 & 6 & 1 & 0\\
		
		47 & Max. temperatures in Melbourne &  & Daily, from from 1981 to 1990  & 3000 & 7 & 0 & 1\\
		
		48 & Min. temperatures in Melbourne &  & Daily, from from 1981 to 1990  & 3000 & 6 & 0 & 1\\
		
		49 & Precipitation in River Hirnant &  & Half-hourly, from Nov. 1, 1972 to Dec. 31, 1972 & 2928 & 6 & 1 & 0\\
		50 & IBM common stock closing prices & & Daily, from Jan. 2, 1962 to Dec. 31, 1965  & 1008 & 10 & 1 & 0\\
		51 & Internet traffic data I &  & Hourly, from Jun. 7, 2005 to Jul. 31, 2005 & 1231 & 10  & 0 & 1\\
		
		52 & Internet traffic data II &  & Hourly, from Nov. 19, 2004 to Jan. 27, 2005 & 1657 & 11 & 1 & 0\\
		
		53 & Internet traffic data III &  & from Nov. 19, 2004 to Jan. 27, 2005 -- Data collected at five minute intervals & 3000 & 6  & 1 & 0\\
		
		54 & Flow of Jokulsa Eystri river &  & Daily, from Jan. 1, 1972 to Dec. 31, 1974 & 1096 & 21  & 0 & 0\\
		
		55 & Flow of O. Brocket &  & Daily, from Jan. 1, 1988 to Dec. 31, 1991 & 1461 & 6  & 1 & 0\\
		
		56 & Flow of Saugeen river I & & Daily, from Jan. 1, 1915 to Dec. 31, 1979  & 1400 & 6 & 0 & 0\\
		
		57 & Flow of Saugeen river II &  & Daily, from Jan. 1, 1988 to Dec. 31, 1991 & 3000 & 30  & 0 & 0\\
		
		58 & Flow of Fisher River & & Daily, from Jan. 1, 1974 to Dec. 31, 1991 & 1461 & 6  & 0 & 1\\
		
		59 & No. of Births in Quebec & & Daily, from Jan. 1, 1977 to Dec. 31, 1990 & 3000 & 6  & 1 & 1\\
		
		60 & Precipitation in O. Brocket & & Daily, from Jan. 1, 1988 to Dec. 31, 1991 & 1461 & 29 & 0 & 0\\
		\midrule
		61 & Min. temperature & \multirow{2}{*}{Porto weather~\cite{cerqueira2017ecml}} &  \multirow{2}{*}{\parbox{5.8cm}{Daily values from Jan. 1, 2010 to Dec. 28, 2013}} & 1456 & 8  & 0 & 1\\
		62 & Max. temperature & & & 1456 & 10 & 0 & 0\\
		\bottomrule    
	\end{tabular}%
	}
	\label{tab:data2}
\end{table}

\end{document}